\documentclass[lettersize,journal]{IEEEtran}
\usepackage{amsmath,amsfonts}
\usepackage{algorithmic}
\usepackage{algorithm}
\usepackage{array}
\usepackage[caption=false,font=normalsize,labelfont=sf,textfont=sf]{subfig}
\usepackage{textcomp}
\usepackage{stfloats}
\usepackage{url}
\usepackage{verbatim}
\usepackage{graphicx}
\usepackage[utf8]{inputenc} 
\usepackage[T1]{fontenc}    
\usepackage{hyperref}       
\usepackage{url}            
\usepackage{booktabs}       
\usepackage{amsfonts}       
\usepackage{nicefrac}       
\usepackage{microtype}      
\usepackage{fancyhdr}       
\graphicspath{{media/}}     
\usepackage[numbers]{natbib}
\usepackage{float}
\usepackage{rotating}
\usepackage{multirow}
\usepackage{arydshln}
\usepackage{subcaption}
\usepackage{placeins}

\begin{document}

\title{SDformerFlow: Spatiotemporal swin spikeformer for event-based optical flow estimation}

\author{Yi Tian and Juan Andrade-Cetto
\thanks{Y. Tian and J. Andrade Cetto are with Institut de Rob\`otica e Inform\`atica Industrial, CSIC-UPC, Barcelona, Spain. emails: ytian@iri.upc.edu, cetto@iri.upc.edu}
\thanks{This work received support from projects EBCON (PID2020-119244GB-I00) and RAADICAL (PLEC2021-007817) funded by MCIN/ AEI/ 10.13039/ 501100011033 and the "European Union NextGenerationEU/PRTR"; the Consolidated Research Group RAIG (2021 SGR 00510) of the Departament de Recerca i Universitats de la Generalitat de Catalunya; and by an FI AGAUR PhD grant to Yi Tian.}
\thanks{Manuscript received DATE; revised DATE. Date of publication DATE.}
\thanks{Digital Object Identifier 10.1109.XXXXXXXXXXXX}
}

\markboth{}%
{Yi \MakeLowercase{\textit{et al.}}: SDformerFlow: Spatiotemporal swin spikeformer for event-based optical flow estimation}


\maketitle

\begin{abstract}
Event cameras generate asynchronous and sparse event streams capturing changes in light intensity. 
They offer significant advantages over conventional frame-based cameras, such as a higher dynamic range and an extremely faster data rate, making them particularly useful in scenarios involving fast motion or challenging lighting conditions. Spiking neural networks (SNNs) share similar asynchronous and sparse characteristics and are well-suited for processing data from event cameras.
Inspired by the potential of transformers and spike-driven transformers (spikeformers) in other computer vision tasks, we propose two solutions for fast and robust optical flow estimation for event cameras: STTFlowNet and SDformerFlow. STTFlowNet adopts a U-shaped artificial neural network (ANN) architecture with spatiotemporal shifted window self-attention (swin) transformer encoders, while SDformerFlow presents its fully spiking counterpart, incorporating swin spikeformer encoders. Furthermore, we present two variants of the spiking version with different neuron models.
Our work is the first to make use of spikeformers for dense optical flow estimation. We conduct end-to-end training for all models using supervised learning. Our results yield state-of-the-art performance among SNN-based event optical flow methods on both the DSEC and MVSEC datasets, and show significant reduction in power consumption compared to the equivalent ANNs. Our code is open-sourced at \href{https://github.com/yitian97/SDformerFlow}{https://github.com/yitian97/SDformerFlow}.
\end{abstract}


\section{Introduction}

\IEEEPARstart{O}{ptical} flow measures pixel motion with photometric consistency in the image plane and is crucial for numerous computer vision and robotics tasks. Whilst traditional frame-based optical flow estimation struggles in low-illumination and fast-motion scenarios, event-based optical flow can better cope with such challenging scenarios thanks to the higher temporal resolution and dynamic range of event cameras. The sparse and asynchronous event streams generated by event cameras directly encode apparent motion patterns, but due to the fundamentally different data throughput of the two camera types, estimating event-based optical flow suggests approaches distinct from those of conventional computer vision.
As with many other computer vision problems, methods using ANNs have demonstrated higher accuracy in event-based optical flow estimation~\cite{Gehrig2021eraft, liu2023tma, Wan22_ieeetip} compared to classical model-based methods~\cite{shiba2022a, paredes2023taming}. 
ANN architectures, however, do not fully exploit the sparse and asynchronous nature of event data, and SNNs have emerged as a promising alternative.
In SNNs, neurons integrate input spike trains and generate a binary spike when the membrane potential reaches a threshold, resetting its value afterward. Neurons are active only when spikes arrive, just as individual event camera pixels are active only when intensity changes. Sharing this event-driven characteristic makes SNNs an energy-efficient option for processing event data. However, directly training deep SNNs is challenging due to the non-differentiability of the spike activity.
The backpropagation through time with surrogate gradient method~\cite{neftci2019surrogate} has bridged neuromorphic computing with the deep learning community, enabling the training of deeper SNNs. Despite this advancement, the performance of SNNs still lags behind that of ANNs for most computer vision tasks.

Is it possible to benefit both from the recent advances in ANN architectures and the spike-driven properties of SNNs to achieve an energy-efficient solution for event-based optical flow estimation with competitive performance? For ANNs, the visual transformer (ViT) and its variant architectures have garnered increasing interest as potential replacements for convolution networks in various computer vision tasks. Due to their inherent locality, convolution-only models struggle to capture temporal correlation and to efficiently represent global spatial dependencies~\cite{dosovitskiy2020ViT, Guizilini2022, Sui2022, huang2022FlowFormer}. At the same time, the self-attention mechanism in ViT architectures can focus on different parts of the input to capture global context effectively.
The integration of ViT, particularly with spatiotemporal attention, has also shown promising results in event-based vision tasks, such as monocular depth estimation~\cite{zhang2022MDE} or action recognition~\cite{Tristan2023EventTransAct, Gao2023EV-ACT}.
Combining SNNs with the ViT architecture for event cameras appears to be a natural choice, as the combination leverages the strengths of both approaches: the temporal dynamics and energy efficiency of SNNs and the representational power of transformers. Moreover, the self-attention mechanism in transformers also shares a biological background with SNNs~\citep{zhou2023spikformericlr, Yao2023attention, yao2023spikedriven, wang2023masked}. 
The spikeformer architecture, the SNN version of the ViT, has been validated mostly on higher level tasks, such as classification~\cite {zhou2023spikformericlr, yao2023spikedriven}, the regression of human pose ~\citep{zou2023sstt}, depth estimation~\citep{zhang2024spikeformerdepth}, video action recognition~\cite{yu2024svformer}, and object detection~\cite{yu2024spikingvit}.

The best-performing event-based optical flow solutions to date use ANNs with correlation volumes~\cite{liu2023tma} or iterative deblurring~\cite{wu2023IDNet}. Correlation volumes require substantial computational and memory resources. Adopting transformers for optical flow in ANNs has also shown superior performance compared to non-transformer-based models, particularly excelling in scenarios involving large displacements due to their ability to capture global dependencies~\cite{huang2022FlowFormer,xu2022gmflow,lu2023transflow,li2023blinkflow,tian2022etflownet}. No one has proposed a pure SNN architecture, specifically utilizing spikeformers for event-based optical flow estimation.

In this work, we study the combination of ViT with SNNs for event-based optical flow estimation. We introduce SDformerFlow, an SNN employing spatiotemporal swin spikeformers. Additionally, for better comparison, we propose STTFlowNet, the ANN counterpart to our SNN model. We conduct end-to-end training using supervised learning. Our work marks the first instance of utilizing spikeformers for optical flow estimation, demonstrating comparable performance to state-of-the-art SNN optical flow estimation methods, while significantly reducing energy consumption. We report on two variants of our SNN model: our first variant SDformerFlow-v1~\cite{tian2024SDformerFlow} and the improved variant SDformerFlow-v2 in this paper with better performance and reduced computational complexity.

Our contributions are threefold: Firstly, we introduce STTFlowNet, a swin transformer-based model for event-based optical flow estimation, equipped with spatiotemporal self-attention to capture dependencies in both the time and space domains. 
Secondly, we present two spiking versions of our architecture, SDformerFlow, with different neuron models, marking the first known utilization of spikeformers for event-based optical flow estimation.
Lastly, we conduct extensive experiments on datasets. Compared with baseline models, our method uncovers the potential of combining transformers with SNNs for regression tasks.

\section{Related work}
\subsection{Learning-based methods for event-based optical flow estimation}

Drawing inspiration from frame-based optical flow techniques, the estimation of event-based optical flow using deep learning has achieved state-of-the-art performance compared to model-based methods~\cite{Gehrig2021eraft,paredes2023taming,shiba2022a}. Early works predominantly employed a U-Net architecture~\cite{zhu2019a,lee2020spikeflownet,hagenaars2021a,ding2021} to predict sparse flow and evaluated it using masks due to limited accuracy where no events are present. Inspired by RAFT flow~\cite{teed2020raft}, Gehrig et al.~\cite{Gehrig2021eraft} proposed E-RAFT and contributed the DSEC dataset and optical flow benchmark~\cite{Gehrig21DSEC}. Since then, methods based on recurrent neural networks with correlation features and iterative refinement strategies have become the state-of-the-art~\cite{Gehrig2021eraft,ding2021,li2023blinkflow}.

Recent studies have shifted their focus towards enhancing the temporal continuity of optical flow estimation, aiming to fully leverage the low latency characteristics of event cameras~\cite{wu2023IDNet, liu2023tma, ponghiran2023efficient-spikeflow}, or integrating richer simulated training datasets~\cite{li2023blinkflow, yang2023ECDDP,luo2024eemflow} to improve accuracy. However, these recurrent refinement methods implicate calculating computationally expensive cost columns and an iterative update scheme that brings latency to the inference phase. Some works explore latency reduction at the expense of slight loss in performance~\cite{wu2023IDNet}.

Another line of work based on SNNs emerges as a computationally efficient solution for event camera optical flow estimation. Most works trained SNNs using self-supervised learning on the MVSEC dataset, yielding sparse flow estimation~\cite{hagenaars2021a, kosta2023adaptivespikenet}. More recent efforts involve training SNNs using supervised learning on the DSEC dataset, resulting in dense flow estimation~\cite{cuadrado2023optical}. 
To incorporate longer temporal correlations into the SNN model, some works utilize adaptive neural dynamics in comparison with event inputs containing richer temporal information~\cite{kosta2023adaptivespikenet}, while others introduce external recurrence~\cite{ponghiran2023efficient-spikeflow}. In~\cite{cuadrado2023optical}, the authors employed 3D convolutions with stateless spiking neurons, neglecting the intrinsic temporal dynamics of the neurons. However, the performance of SNNs still falls behind that of ANNs. 
While some ANN methods incorporate transformer architectures in some of their stages \cite{huang2022FlowFormer,xu2022gmflow,lu2023transflow,li2023blinkflow,tian2022etflownet} and show performance improvements, no one has ever combined SNNs with transformer architectures for optical flow estimation.

\subsection{Spikeformer}

Recently, the combination of SNNs and transformer architectures has garnered increasing interest in the neuromorphic community~\cite{zhou2023spikformericlr,Yao2023attention, yao2024metaformer,zhou2024qkformer}. Zhou et al.~\cite{zhou2023spikformericlr} initially proposed spiking self-attention, which eliminates the softmax function as the spike-formed query and key naturally maintains non-negativity. Building upon this, Yao et al.~\cite{yao2023spikedriven} introduced a fully spike-driven transformer with spike-driven self-attention, leveraging only mask and addition operations to facilitate hardware implementation. Later, they expanded it into a meta-architecture~\cite{yao2023spikedriven} for classification, detection, and segmentation tasks. Shi et al.\cite{Shi2024spikingresformer} pointed out that the previous spikeformers rely on a shallow convolutional network for feature extraction and lack proper scaling methods~\cite{zhou2023spikformericlr,yao2023spikedriven}. They proposed a multi-stage architecture with a dual spike self-attention (DSSA) and a proper scaling method to address the problem. More recently, Zhou et al.\cite{zhou2024qkformer} proposed QKFormer with a Q-K attention that adopts spike-based components for the query and key with linear complexity.
While most spikeformers only apply spatial-wise attention in a single time step~\cite{zhou2023spikformericlr, yao2023spikedriven, Shi2024spikingresformer, zhou2024qkformer}, some works also incorporate spatiotemporal attention~\cite{zou2023sstt, wang2023STSA,yu2024svformer}. However, none of the previous works have utilized the swin variant of the spikeformer for optical flow estimation.
\section{Method}

\subsection{Preliminaries}

\subsubsection{Spiking neurons}
Spiking neurons are the fundamental units of SNNs. Unlike conventional ANNs that use continuous activation functions, spiking neurons communicate through discrete spikes known as action potentials. The leaky integrate-and-fire (LIF) model is the most commonly used neural model in the literature, although numerous recent studies have investigated adaptive neurons to achieve better performance \cite{kosta2023adaptivespikenet, yao2022glif}.
\paragraph{Leaky Integrate-and-Fire}
The LIF model is widely adopted in the literature due to its simplicity of implementation and low computational cost.
In SDformerFlow-v1, we use the Spikejelly~\cite{SpikingJelly} implementation of the LIF neuron model for all layers, and set $V_{th} = 0.1$ and $\tau_{m} = 2$.

The dynamics of LIF neurons at time step $t$ with hard reset can be modeled as
\begin{align}
   &H[t] = V[t-1] + \frac{1}{\tau}(X[t]-(V[t-1]-V_{reset}))\\
    &S[t] = \Theta(H[t]-V_{th})\\
    &V[t] = H[t](1-S[t])+V_{reset}S[t] \;,
\end{align}
where $X[t]$ represents the inputs at time step $t$. $H[t]$ is the membrane potential of the LIF neuron before the neuron fires, while $V[t]$ is the membrane potential after it fires. $H[t]$ changes according to the presynaptic spikes received, and $\tau$ is the time constant. When $H[t]$ reaches the threshold $V_{th}$, the neuron fires, generating a spike $S[t]$, and $V[t]$ resets to $V_{reset}$. $\Theta(V)$ is the Heaviside step function, which outputs a zero value for negative arguments and one for positive arguments.

\paragraph{Parallel Spiking Neuron (PSN)}
Fang et al.~\cite{fang2023psn} proposed PSN, which enables parallelizable neuronal dynamics after removing the resetting mechanism. It maximizes the utilization of temporal information and yields extremely high simulation speed. In SDformerFlow-v2, we change to the use of PSN neurons with learnable parameters.

In PSN, the neuron dynamics for LIF without reset mechanism become
\begin{align}
H[t] &= \sum_{i=0}^{T-1} W_{t,i} \cdot X[I] \\
W_{t,i} &= \frac{1}{\tau} \left(1 - \frac{1}{\tau}\right)^{t-i} \cdot \Theta(t - i) \;,
\end{align}
where $W_{t,i} $ is the weight between input $X[i]$ and membrane potential $H[t]$. The non-iterative formulation enables to parallelize the neuron state across time steps with a learnable weight matrix ${\bf W}$ and a learnable threshold vector ${\bf B}$:
\begin{align}
\mathbf{H} &= \mathbf{WX}, & \bf W &\in \mathbb{R}^{T \times T}, \mathbf{X} \in \mathbb{R}^{T \times N} \nonumber \\
\mathbf{S} &= \Theta(\mathbf{H} - \mathbf{B}), & \mathbf{B} &\in \mathbb{R}^{T}, \mathbf{S} \in \{0,1\}^{T \times N} \nonumber
\end{align}
where $T$ are time steps and $N$ is batch size. In this way, the neuron state integrates the information from all time steps and avoids the iterative process. 

\subsubsection{Surrogate gradient (SG)}
 One of the key challenges in training SNNs has been the non-differentiable nature of spike generation, which precludes the use of standard gradient-based optimization commonly employed in traditional ANNs. The SG method~\cite{neftci2019surrogate} has bridged the gap by approximating the non-differentiable spike function with a continuous, differentiable surrogate during the backpropagation phase. The common surrogate function choices include inverse tangent and sigmoid.

\subsubsection{Spike Self-Attention (SSA)}
The self-attention in ANN is composed of three floating-point components: query ($Q$), key ($K$), and value ($V$). Zhou et al. \cite{zhou2023spikformericlr} first proposed SSA which is based on spike-forms for $Q$, $K$, and $V$,
\begin{align}
&Q_s, K_s, V_s = SN (BN (Linear(I))) \\
&SSA'(Q_s, K_s, V_s) = SN \left(QK^T V * s\right) \\
&\text{\small$SSA(Q, K, V) = SN \left(BN\left(Linear(SSA'(Q_s, K_s, V_s))\right)\right)$},
\end{align}
where $Q_s, K_s, V_s$ are the spikes form of query, key, and value. $I$ denotes the input. $Linear$, $BN$, and $SN$ denote the linear layer, batch normalization layer, and spiking neuron, respectively, and $s$ is a scaling factor used to control large output values to avoid gradient vanishing.

\subsection{Event Input Representation}
We divide the event stream into non-overlapping chunks according to the optical flow ground truth rate. Each chunk, comprising $N$ events within a fixed time window, is represented as $ E = \{(x_i,y_i,t_i,p_i)\}_{i\in[N]} $, where $t_i$ is the timestamp and $p_i$ denotes polarity. We preprocess each event chunk into an event discretized volume representation $\bf V$ using a set of $B$ bins, following the methodology introduced in~\cite{zhu2019a},
\begin{equation}
   {\bf  V}(x,y,t) = \sum_i p_i\kappa(x-x_i) \kappa(y-y_i)\kappa(t-t_i) \;.
\end{equation}
Timestamps are normalized and scaled to the range $[0,B-1]$, $t_i =(B-1)(t_i-t_0)/(t_N-t_1)$; and $\kappa(a)=\max(0,1-|a|)$ is a bilinear sampling kernel.
We encode spatiotemporal information into channels to enable the neural network to learn large temporal correlations. For the ANN model, we take the previous and current chunks of event voxels, dividing the total temporal channels into $n$ blocks. Each event input block comprises ${2B}/{n}\times H\times W$ bins. In our case, $n =2$.

\begin{figure} [!t]
    \centering
	\includegraphics[clip, trim=17cm 16cm 25cm 15cm,width = 0.75\columnwidth]{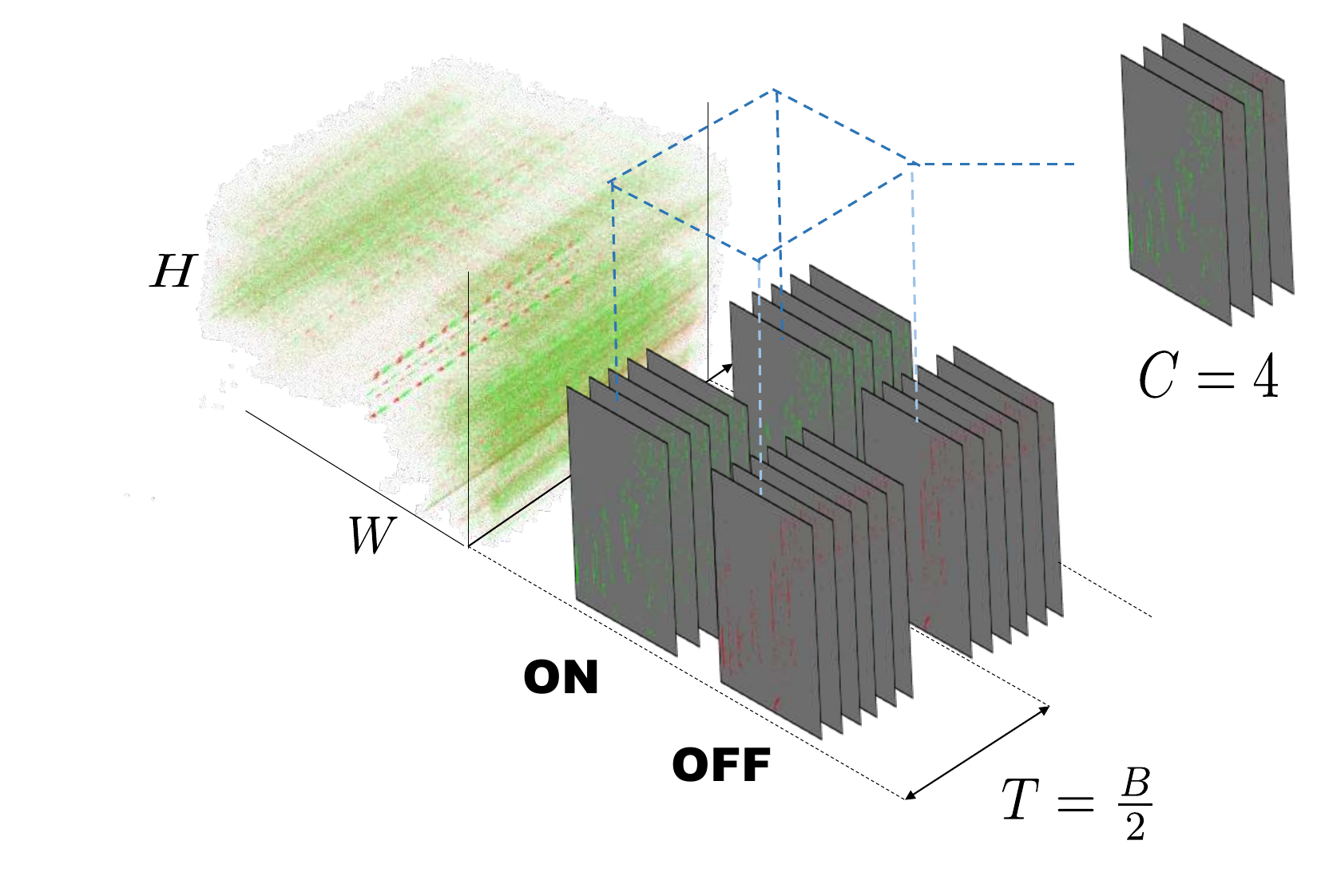}
	\caption{Event input representation for SDformerFlow.}
 \label{fig:input}
\end{figure}

For the SNN model, to mitigate the computational burden associated with large time steps, we use only one event voxel chunk. Similarly, we partition the temporal channel, containing $B$ bins, into $n$ blocks along with their corresponding polarities $p$. This yields an event representation of size $T \times 2n \times H\times W$, with $T = {B}/{n}$  time steps. In most of our implementation, we set $B=10$ and $n = 2$. This representation aligns with the spike representation outlined in~\cite{kosta2023adaptivespikenet,lee2020spikeflownet}. Each event chunk comprises $C = 4$ channels and $T = {B}/{2}$ time steps, as illustrated in Fig.~\ref{fig:input}.

\subsection{Network Architecture}

\begin{figure*}[!t]
\begin{center}
\includegraphics[clip, trim=0cm 2cm 0cm 0cm,width=\textwidth]{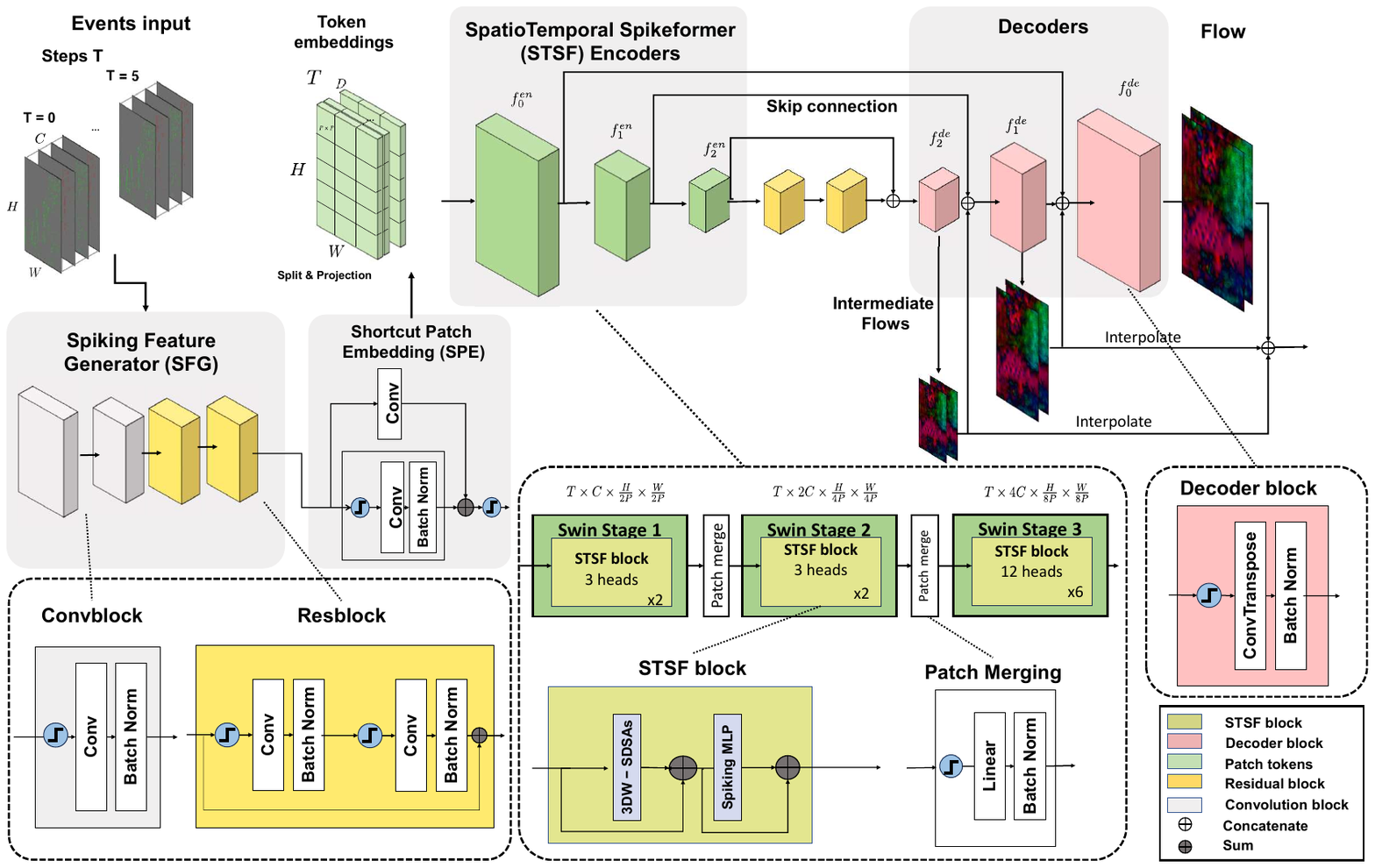}
\end{center}
 \vspace{-0.5cm}
   \caption{SDformerFlowNet-v2 architecture.}
\label{fig:architecture}
\end{figure*}

\begin{figure}[t]
\begin{center}
\includegraphics[clip, trim=2cm 15.5cm 18cm 0cm,width=0.7\columnwidth]{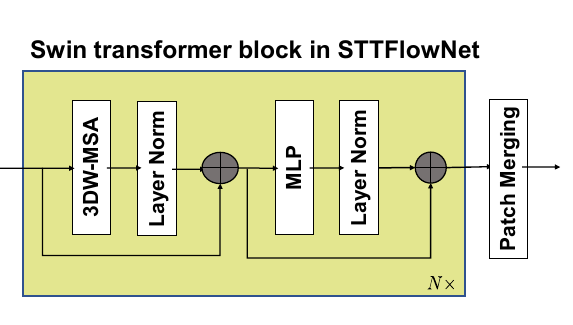}
\end{center}
   \caption{Spatio-temporal swin transformer blocks in STTFlowNet.}
\label{fig:stt-swin-architecture}
\end{figure}

The network architecture pipelines of our proposed SNN methods SDformerFlow (Fig.~\ref{fig:architecture}) and its ANN equivalent STTFlowNet are similar. We adopt an encoder-decoder architecture, widely utilized in event-based optical flow literature~\cite{hagenaars2021a,zhu2019a,zhu2018a,cuadrado2023optical, luo2024eemflow}. For STTFlowNet, the architecture of the swin transformer blocks resembles that of~\cite{liu2022video}. As shown in Fig. \ref{fig:stt-swin-architecture}, each swin block contains a 3D window multi-head self-attention (3DW-MSA) module, followed by a module consisting of two multi-layer-perceptron (MLP) blocks. Layer normalization (LN) is applied after each module, incorporating residual connections. In the 3DW-MSA module, unlike in the original video swin transformer implementations~\cite{liu2022video}, we utilize scaled cosine attention and logarithmic continuous relative position bias (CPB) from swin transformer v2~\cite{liu2022swinv2} to enhance the model's scaling capability. 
In the following sections, we focus on detailing the architecture of our SNN model: SDformerFlow.

For SDformerFlow, the primary architecture comprises three parts:
a) spiking feature generator (SFG) with shortcut patch embedding (SPE), 
b) spatiotemporal swin spikeformer (STSF) encoders, and 
c) spike decoders and flow prediction.
The event stream initially enters the SFG module, which outputs spatiotemporal embeddings for the STSF encoders, which in turn generate spatiotemporal features hierarchically. Subsequently, the output from each encoder is concatenated to the decoder at the same scale to predict the flow map. Two additional residual blocks exist between the encoder and decoder modules.

\begin{figure}[!t]
\begin{center}
\includegraphics[clip, trim=6cm 9cm 12cm 2cm,width=0.7\columnwidth]{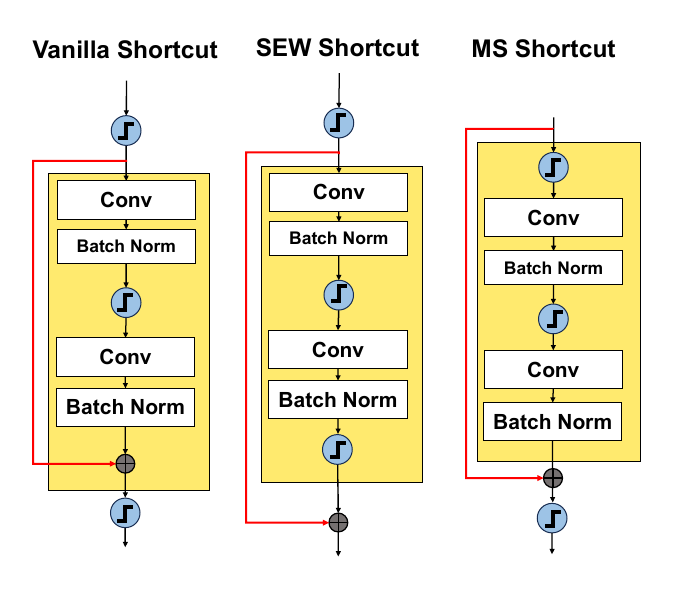}
\end{center}
 \vspace{-0.5cm}
   \caption{Shortcuts in SNN. The left and middle blocks illustrate the vanilla and spike-element-wise (SEW) shortcuts commonly used in other architectures~\cite{zhou2023spikformericlr}. The right block shows the membrane potential (MS) shortcut used in our SDformerFlow.}
\label{fig:shortcuts}
\end{figure}

In previous spikeformer implementations~\cite{zhou2023spikformericlr}, residual shortcuts utilize either vanilla or spike-element-wise shortcuts (SEW)~\cite{Fang2021SEW}. Conversely, in SDformerFlow, we opt for using membrane-potential shortcuts (MS)~\cite{Hu2024MS_res}. 
The vanilla shortcut adds spikes into the memory potential values, which cannot achieve identity mapping and show degradation problems. In SEW shortcuts, the residuals are applied after the spikes, which results in undesirable integration, whereas with MS shortcuts, residuals are applied before the spikes to preserve the spike-driven property. Fig.~\ref{fig:shortcuts} illustrates the main differences between vanilla shortcuts, SEW shortcuts, and MS shortcuts. 

Specific implementation details of each main block in the SDformerFlow architecture are:

\subsubsection{Spiking Feature Generator with Shortcut Patch Embedding}
It comprises two stages: an SFG block to generate spatiotemporal features, followed by an SPE block to project them into token embeddings for the STSF encoder module.

\paragraph{Spiking Feature Generator} In the first stage, we process the event input through a spiking convolutional module {\em Conv}$(\cdot)$ followed by two residual blocks with MS shortcut {\em MSRes}$(\cdot)$ to downsample the resolution by half. This projection results in a feature map of shape $T \times C \times {H}/{2} \times {W}/{2}$.

Given the events input $I$, the feature generator module can be formulated as
\begin{align}
    &\hat{z} = BN(Conv(SN(Conv_{head}(I))))\\
    &z = MSRes(\hat{z})\\
   &\text{\footnotesize$MSRes(\hat{z}) =\hat{z} + BN_2(Conv_2(SN_2(BN_1(Conv_1(SN_1(\hat{z})))))$} \;,
\end{align}
where $BN$ and $SN$ again account for batch normalization and spiking activation, respectively.  
For STTFlowNet, both the former and latter chunks are fed into a shared Resblock module while retaining the spatial dimension.

\paragraph{Shortcut Patch Embedding}
In the second stage, we split the feature map into spatial patches of size $P \times P$, maintaining the time steps as the temporal dimension. This operation creates spatiotemporal tokens of size $1 \times P \times P$, projecting the spatial-temporal features into spike embeddings of shape $T \times C \times {H}/{(2P)} \times {W}/{(2P)}$. Inspired by~\cite{zhou2024qkformer}, we add a deformed shortcut for the patch embedding module, which boosts the performance. A convolutional layer $Conv_{deformed}$ with kernel size of $1 \times 1$ and stride size of 2 is applied to the residual to meet the output shape of the embeddings. The SPE block can be formulated as
\begin{align}
&z_{res} = Conv_{deformed}(I)\\
&z = BN(Conv(SN(I))) +z_{res} \;.
\end{align}

\subsubsection{Spatiotemporal Swin Spikeformer Encoder}

The STSF module draws inspiration from the video swin transformer~\cite{liu2022video} and the recent spikeformers~\cite{zhou2023spikformericlr,yao2023spikedriven,zhou2024qkformer}. Its detailed architecture is illustrated in Fig.~\ref{fig:architecture}.

We adopt four stages of swin transformers, with each stage comprising $2-2-6-2$ numbers of STSF blocks successively, followed by a spiking patch merging layer to reduce the dimension by half. 

Within the same swin layer, the window-based multi-head attention in the first block is denoted as 3DW-SDSA, as the regular window partitioning is performed. In the latter blocks, the window is shifted along the three axes following the same practices as in \cite{liu2021swin,liu2022video}, which is denoted as 3DSW-SDSA. The STSF block can be formulated as
\begin{align}
    &\hat{z}^m =3DW-SDSA(z^{m-1}) + z^{m-1}\\
    &z^m = SMLP( \hat{z}^m)) +  \hat{z}^m\\
     &\hat{z}^{m+1} =3DSW-SDSA(z^m) + z^m\\
     &z^{m+1} = SMLP(\hat{z}^{m+1}) + \hat{z}^{l+m}\\
     &\text{\footnotesize$SMLP(z) = BN_2(Linear_2(SN_2(BN_1(Linear_1(SN_1(z))))))$}, 
\end{align}
where  $\hat{z}^l$ are the output features of the spiking 3DW-SDSA module or 3DSW-SDSA, and $z^l$ is the output for the spiking MLP module for block $l$.

\begin{figure}[!t]
\centering
\includegraphics[clip, trim=0cm 9cm 10cm 0cm,width=\columnwidth]{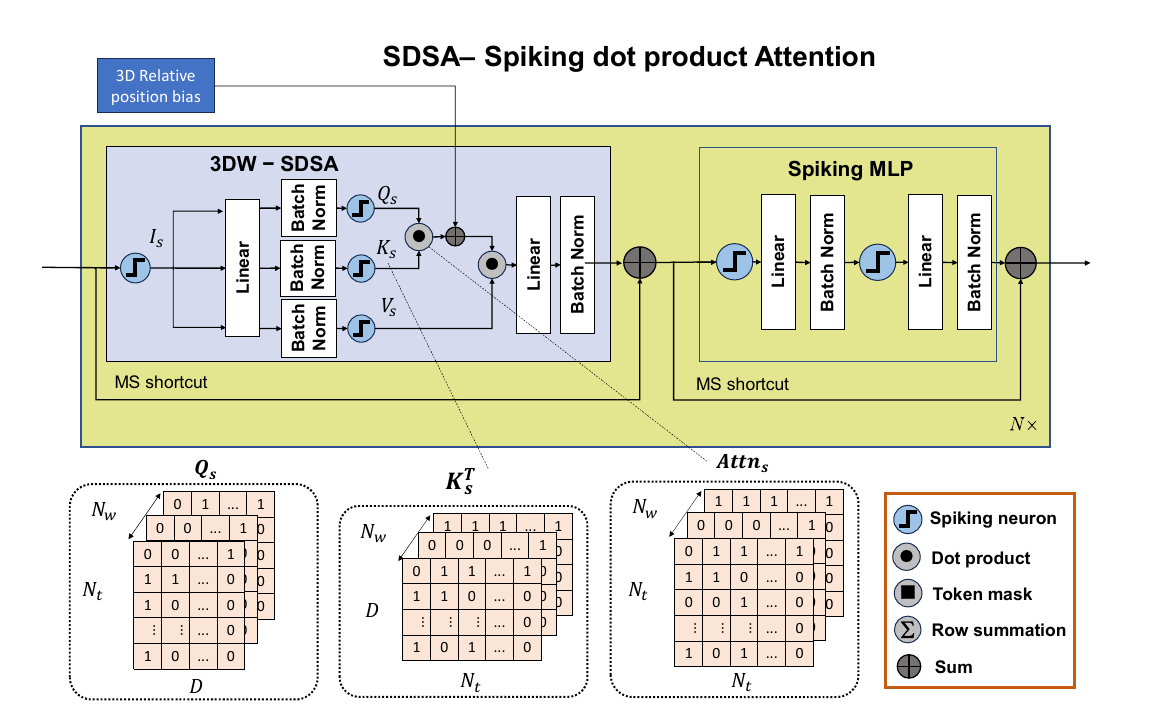}
 \vspace{-0.5cm}
   \caption{Spike-driven self-attention (SDSA) block with spiking dot product attention utilized in SDformerFlow-v1. }
   \label{fig:swinblocks-sdsa}
\end{figure}

\begin{figure}[t]
    \centering
    \includegraphics[clip, trim=0cm 9cm 10cm 0cm,width=\columnwidth]{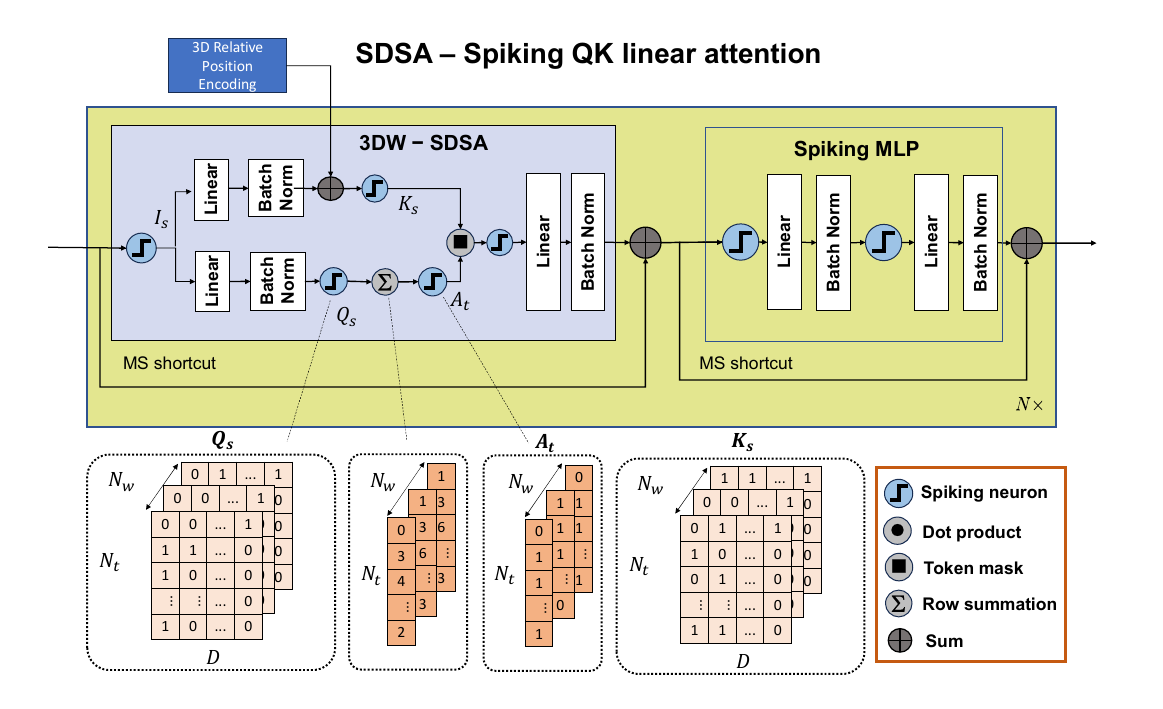}
    \vspace{-0.5cm}
\caption{Spike-driven self-attention (SDSA) block with spiking QK linear attention utilized in SDformerFlow-v2.}
\label{fig:swinblocks-qk}
\end{figure}

Each STSF block comprises a spiking multi-head spiking driven self-attention (SDSA) block with a 3D shifted window (3DW), followed by a spiking MLP block (see Fig.~\ref{fig:swinblocks-sdsa}). Each spatiotemporal token of shape $T \times H \times W$ is partitioned into non-overlapping 3D windows of size $T_w \times H_w \times W_w$. We employ a window size of $2 \times 9 \times 9$ for cropped resolution and $2 \times 15 \times 15$ when fine-tuning the model on a full resolution of $480 \times 640$. The SDSA is performed within the window.
We utilize different numbers of attention heads $3, 6, 12,24$ for the STSF blocks in different stages. We have implemented two types of spiking self-attention that we called spiking dot product attention (Fig.\ref{fig:swinblocks-sdsa}) and spiking QK linear attention (Fig.\ref{fig:swinblocks-qk}). The details of the SDSA modules are explained as follows:

\paragraph{SDSA - Spiking dot product attention}
In our SDSA - spiking dot product attention block, the query, key, and value tensors, denoted as $Q_s, K_s, V_s$, are spiking tensors. We use dot product attention, and since the attention maps are naturally non-negative, softmax is unnecessary~\cite{zhou2023spikformericlr}. We apply a scale factor $s$ for normalization to prevent gradient vanishing in the case of using LIF. Note that for adaptive neurons like GLIF~\cite{yao2022glif} and PSN~\cite{fang2023psn} no scaling is needed since the threshold values can be learned. The single-head SDSA can be formalized as
\begin{equation}
  \text{\small$SDSA(Q_s,K_s,V_s) = BN(Linear((Q_sK_s^T+PE)V_s))  $} \;.
\end{equation}

\paragraph{SDSA - Spiking QK linear attention}
The spiking dot product attention has a computational complexity of $O(N_w*N_t^2*D)$. In addition, adding the 3D positioning bias into the spike attention map introduces floating-point computations into the attention. Inspired by recent work to relax the computational complexities in spiking self-attention~\cite{yao2023spikedriven,zhou2024qkformer}, we adapted QK token attention~\cite{zhou2024qkformer} into our 3D window attention. Given the spiking input $I$, we can obtain the spike form $Q_s, K_s \in \mathbb{R}^{T \times N_w \times W_h \times W_w \times D}$, where $T,N_w,W_h,W_w,D$ denote time steps, number of windows, window height, window width, and hidden dimensions, respectively. We add the positioning encoding (PE) parameters into the states of $k$ before the spiking activation. In the case of single-head attention, $Q_s, K_s$ can be reshaped into $N_w \times N_t \times D$, where $N_t=T*W_h*W_w$ is the number of tokens that includes the spatial and temporal dimensions of one window. For the spatiotemporal QK attention part, first, we generate the spiking token attention vector $A_t$ by summing the dimensions of $Q_s$ matrix followed by a spiking neuron, which models the importance of different tokens. Next, the attention output $z'$ is obtained by applying Hadamard product $\otimes$ between the token attention vector $A_t$ and $K_s$. This is equivalent to applying a token (column) mask operation to $K_s$. In such a way, we achieve linear complexity attention with $O(N_w*D)$ while preserving the spike-driven properties. Finally, a spiking neuron is applied after the attention output, followed by a linear projection layer. The detail of the attention process can be formulated as 
\begin{align}
&\text{\small$Q_s = SN(BN(Linear (I_s))), \quad Q_s \in \mathbb{R}^{N_w \times N_t \times D}$}\\
&\text{\small$K_s = SN(BN(Linear (I_s)) + PE), \quad K_s \in \mathbb{R}^{N_w \times N_t \times D}$}\\
&A_t = SN ( \sum_{i=0}^{D} Q_s^{i,j} ), \quad   A_t \in \mathbb{R}^{N_w \times N_t \times 1} \\
&z'= A_t \otimes K_s, \quad z  \in \mathbb{R}^{N_w \times N_t \times D} \\
&z = BN (Linear(SN(z'))) \;.
\end{align}

\paragraph{Spiking Patch Merge (SPM)}
An SPM layer is added after each STSF encoder except for the last one. It comprises a linear layer followed by a batch normalization layer to downsample the feature map in the spatial domain while maintaining the temporal dimension. The SPM layer is implemented with
\begin{gather}
    SPM(z) = BN(Linear(SN(z))) \;.
\end{gather}

\subsubsection{Spike Decoder Block}
The decoder consists of three transposed convolutional layers {\em ConvTrans}$(\cdot)$, each increasing the spatial resolution by a factor of two. A skip connection from each STSF encoder is concatenated to the prediction output from the corresponding decoder of the same scale. Flow prediction is generated at each scale and concatenated to the decoders. Loss is applied to the flow prediction upsampled to the full resolution.
Given the output from the STSF of each scale $l$ as $z_{en}^l$, the output of each decoder $z_{de}^l$ and the flow prediction can be formulated as 
\begin{align}
    &z_{de}^l = BN(ConvTrans(SN(z_{en}^l \oplus pred(z_{de}^{l-1})))) \\
    &pred(z_{de}^l) = Conv(z_{de}^l) \;.
\end{align}

\subsection{Loss Function}
We train our model with supervised learning using the mean absolute error between the predicted optical flow $\mathbf{u}^{pred}_i=(u^{pred}_i, v^{pred}_i)$ and the ground-truth flow $\mathbf{u}^{gt}_i=(u^{gt}_i, v^{gt}_i)$. Our loss function can be formulated as
\begin{gather}
L = \frac{1}{n} \sum_{i=1}^n |\mathbf{u}^{pred}_i-\mathbf{u}^{gt}_i|
\end{gather}
where $n$ is the number of valid ground truth pixels. For SNN, we employ surrogate gradient~\cite{neftci2019surrogate} with backpropagation through time to train the network. We use the inverse tangent as the surrogate function with a width of $2$.

\section{Experiments}
\subsection{Dataset and training details}

First, we use the DSEC dataset~\cite{Gehrig21DSEC} for both training and evaluation. The DSEC dataset is a comprehensive outdoor stereo event camera dataset featuring a resolution of $640 \times 480$. Ground-truth optical flow annotations are provided at a rate of 10Hz for some of the sequences. To address the lack of ground truth in the test set, we adopt a similar data split strategy as in~\cite{cuadrado2023optical}, dividing the training sequences into training and validation sets. Notably, we exclusively use rectified event data from the left camera.
During training and validation, we perform data augmentation techniques, including random horizontal and vertical flips, as well as random crops on a $288 \times 384$ resolution.

We train the models on three NVIDIA GeForce RTX 2080 Ti GPUs and employ the AdamW optimizer for a total of 80 epochs, ensuring convergence. The initial learning rate is set to 0.001 with a weight decay of 0.01. Additionally, we implement a multistep scheduler that halves the learning rate every 10 epochs.
To mitigate performance degradation when scaling up to full resolution, we conduct fine-tuning on the full-resolution data for an additional 30 epochs before testing. Given the constraints of GPU memory, training at full resolution requires a reduced batch size (1 or 2). During the evaluation test, we disable the tracking of running states for batch normalization layers.

\begin{figure*}[!t]
\begin{tabular}
{c@{\hspace{1mm}}c@{\hspace{1mm}}c@{\hspace{1mm}}c@{\hspace{1mm}}c@{\hspace{1mm}}c}
\includegraphics[width = 3.4cm]{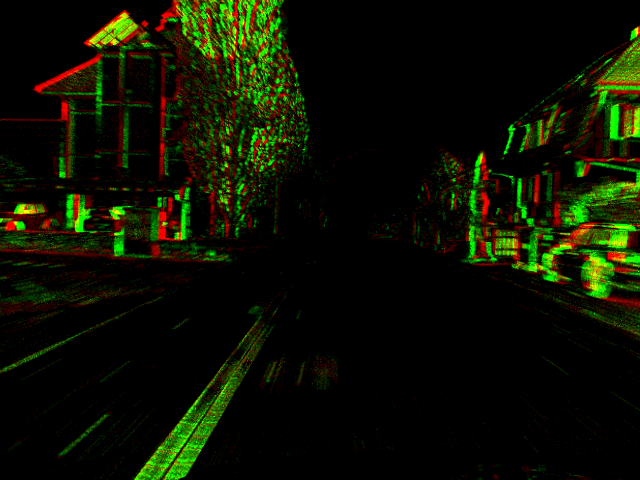}&
\includegraphics[width = 3.4cm]{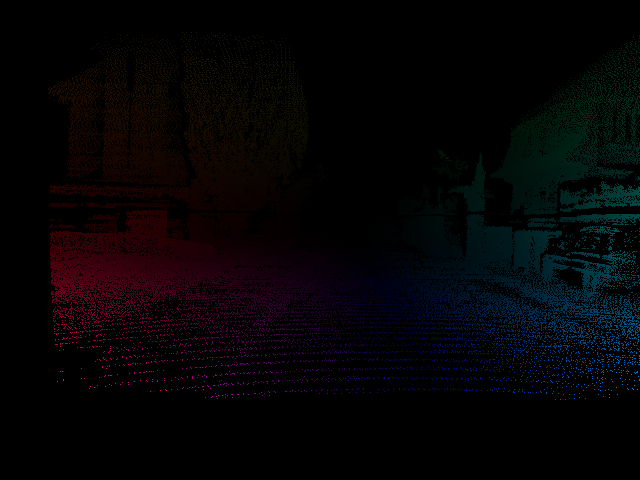}&
\includegraphics[width = 3.4cm]{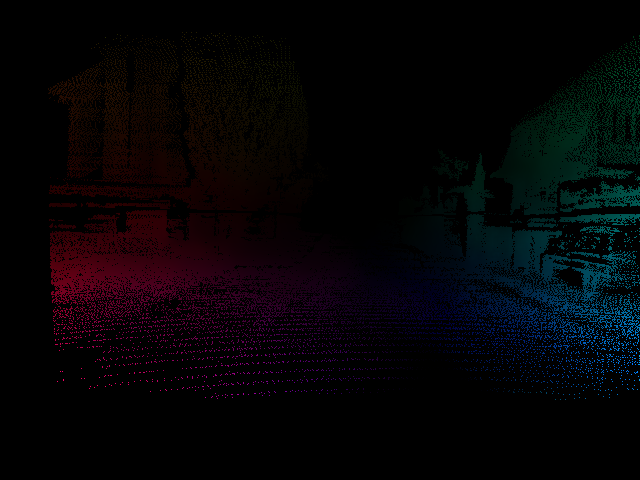}&
\includegraphics[width = 3.4cm]{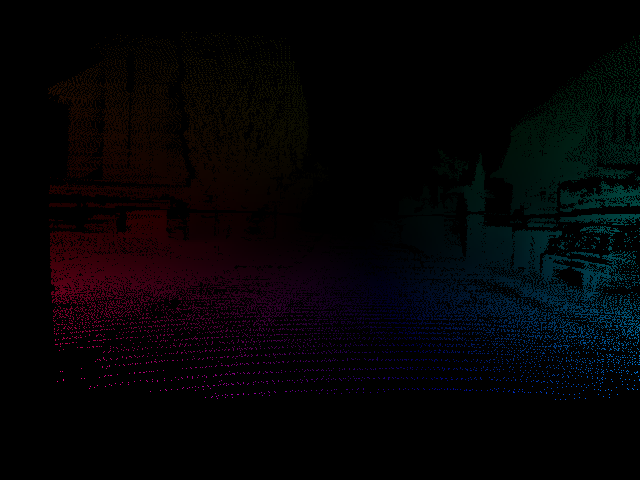}&
\includegraphics[width = 3.4cm]{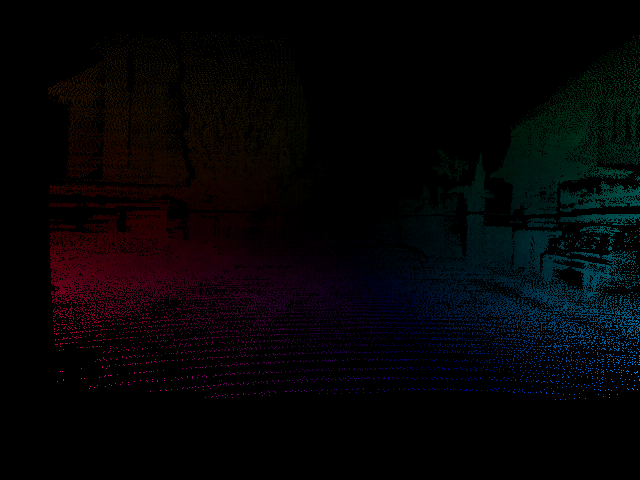}\\
\includegraphics[width = 3.4cm]{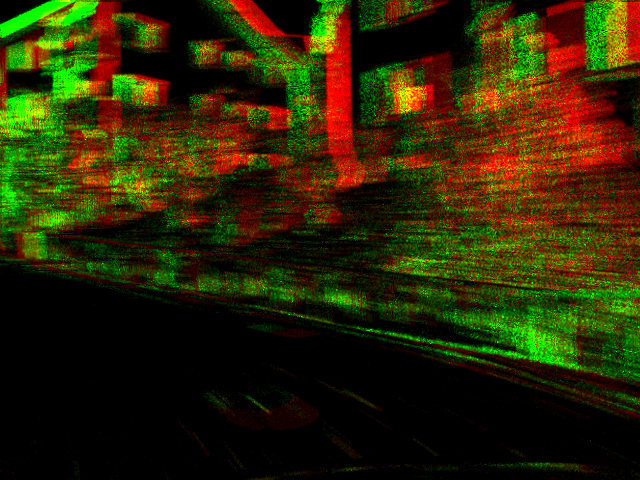}&
\includegraphics[width = 3.4cm]{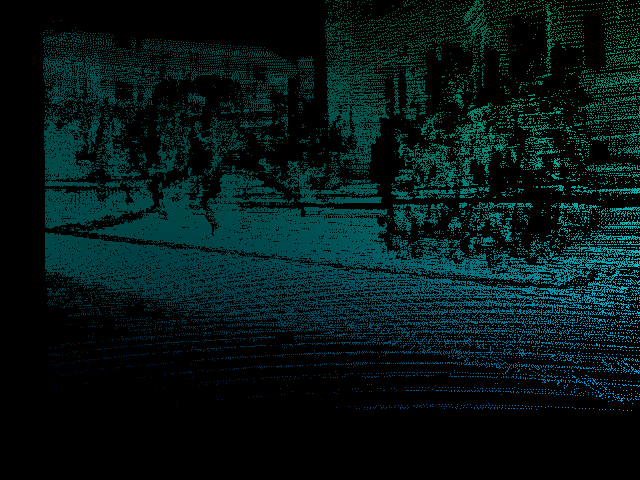}&
\includegraphics[width = 3.4cm]{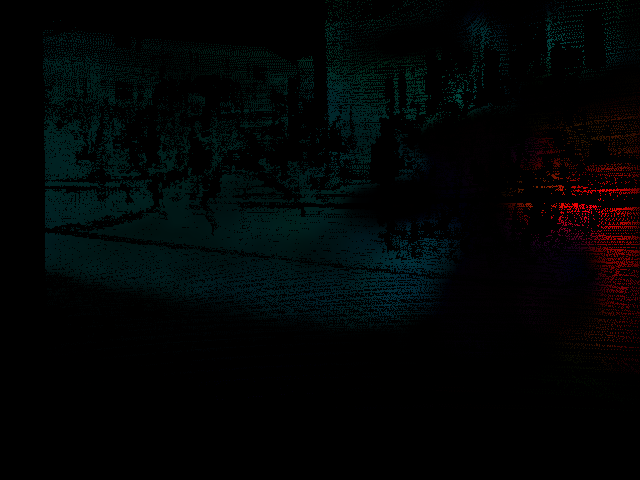}&
\includegraphics[width = 3.4cm]{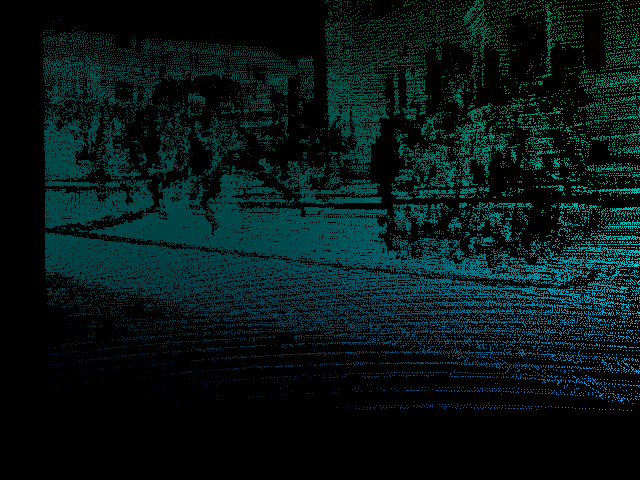}&
\includegraphics[width = 3.4cm]{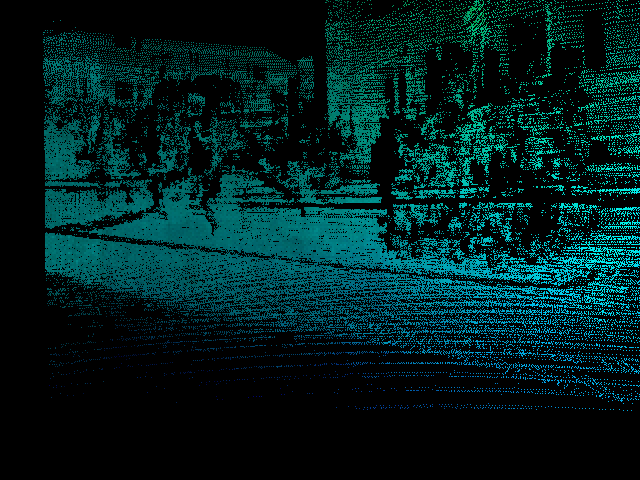}\\
\includegraphics[width = 3.4cm]{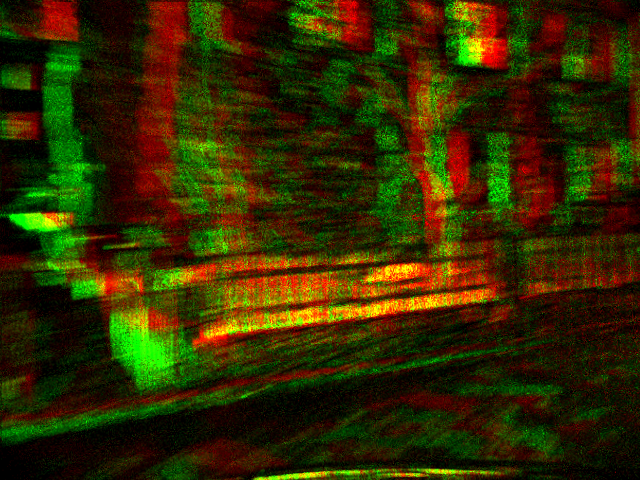}&
\includegraphics[width = 3.4cm]{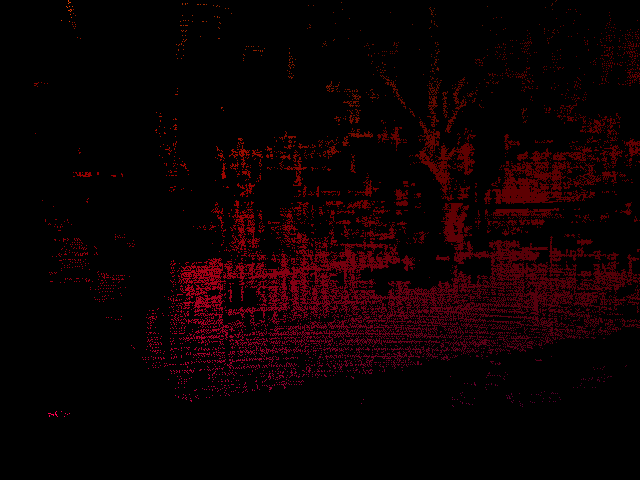}&
\includegraphics[width = 3.4cm]{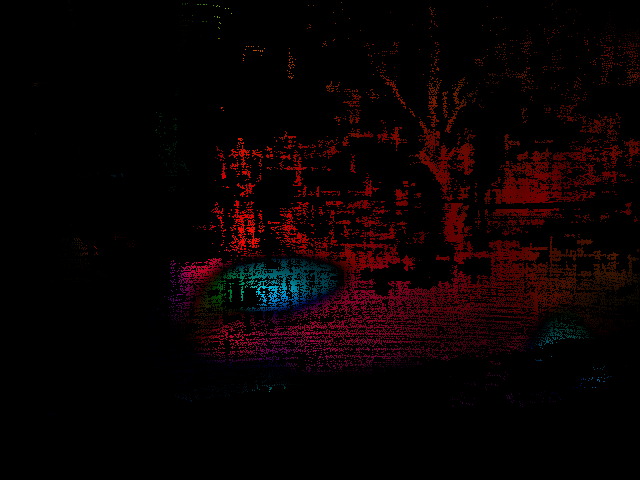}&
\includegraphics[width = 3.4cm]{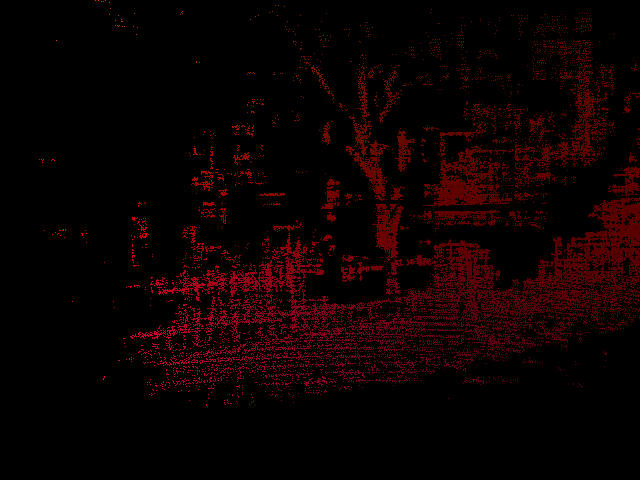}&
\includegraphics[width = 3.4cm]{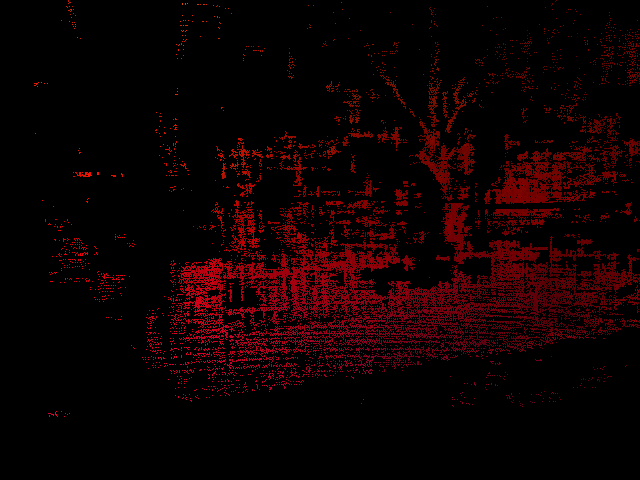}\\

{\footnotesize (a) events} & {\footnotesize (b) ground truth}&{\footnotesize (c) EVFlowNet \cite{zhu2019a}} &{\footnotesize (d) STTFlowNet (Ours)} &{\footnotesize (e) SDformerFlow-v1 (Ours)} \vspace{0.2cm}
\end{tabular}
\caption{Qualitative results for optical flow evaluated on the DSEC validation subset. The first column presents the event input, and the second column shows the ground truth optical flow (For visualization we masked estimated flow where ground truth flows are available). EVFlowNet is our baseline ANN method (best viewed in color).}
\label{fig:DSEC_qualitative}
\end{figure*}

\begin{figure*}[!t]
\begin{tabular}{c@{\hspace{1mm}}c@{\hspace{1mm}}c@{\hspace{1mm}}c@{\hspace{1mm}}c}
\includegraphics[width = 3.4cm]{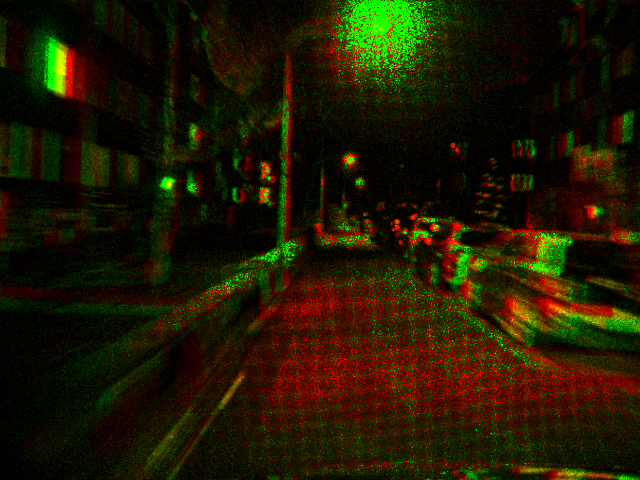}&
\includegraphics[width = 3.4cm]{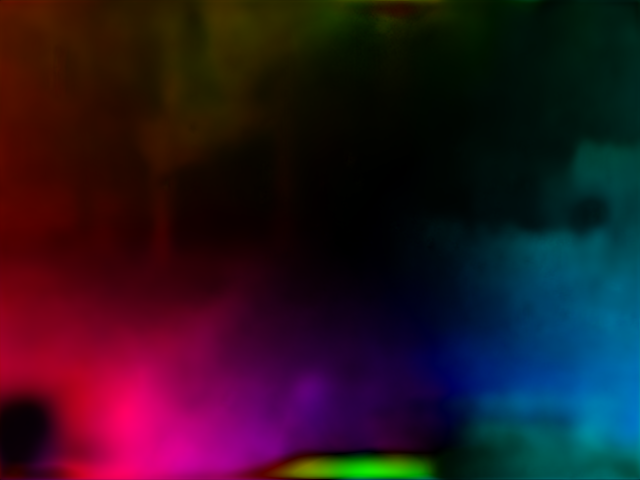}&
\includegraphics[width = 3.4cm]{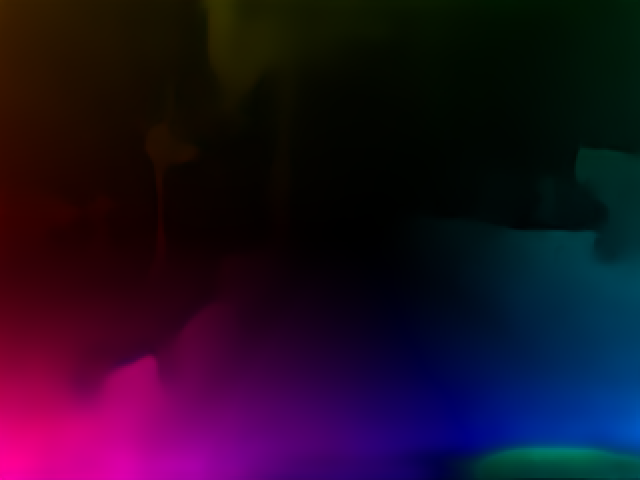}&
\includegraphics[width = 3.4cm]{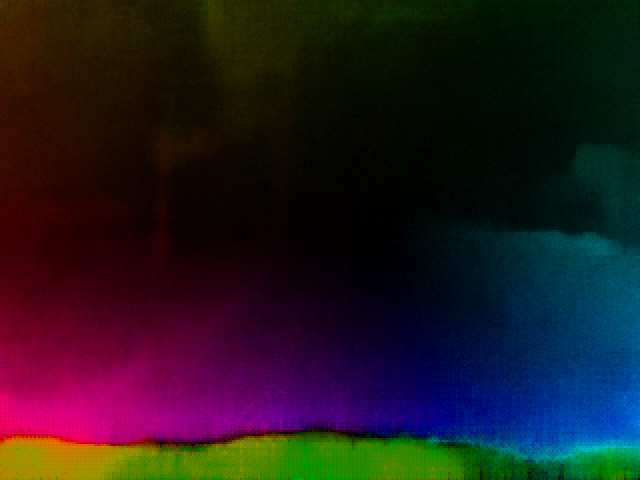}&
\includegraphics[width = 3.4cm]{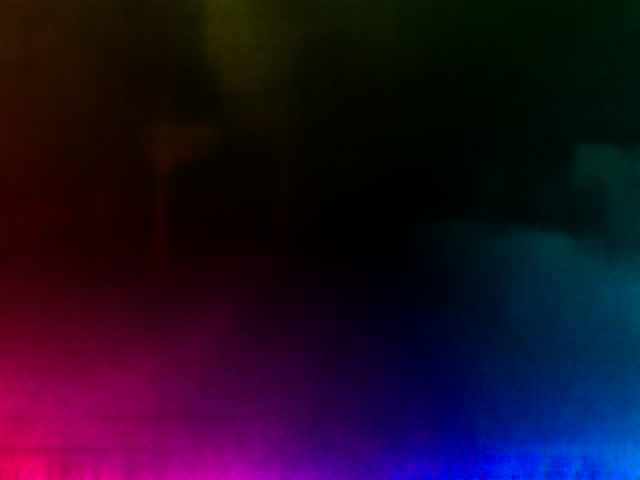}\\
\includegraphics[width = 3.4cm]{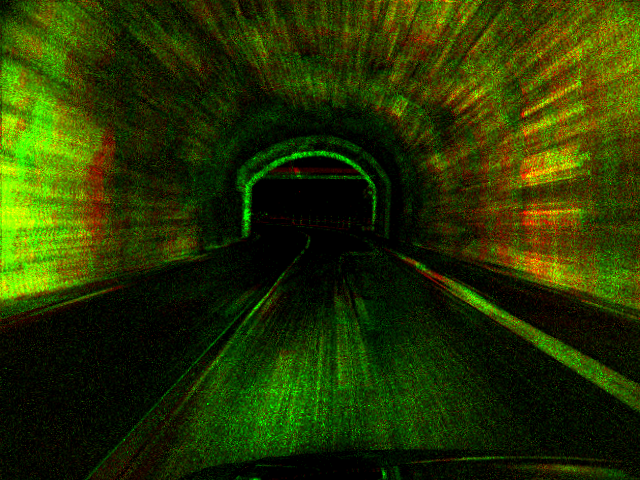}&
\includegraphics[width = 3.4cm]{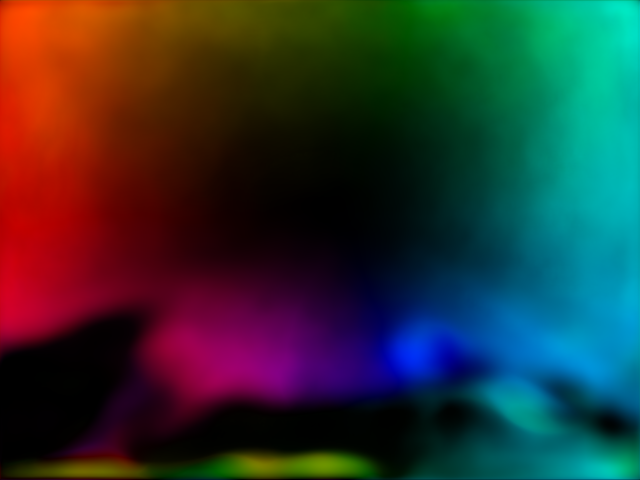}&
\includegraphics[width = 3.4cm]{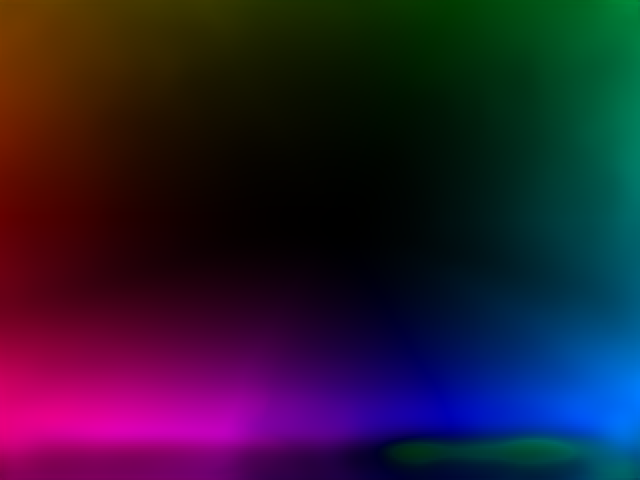}&
\includegraphics[width = 3.4cm]{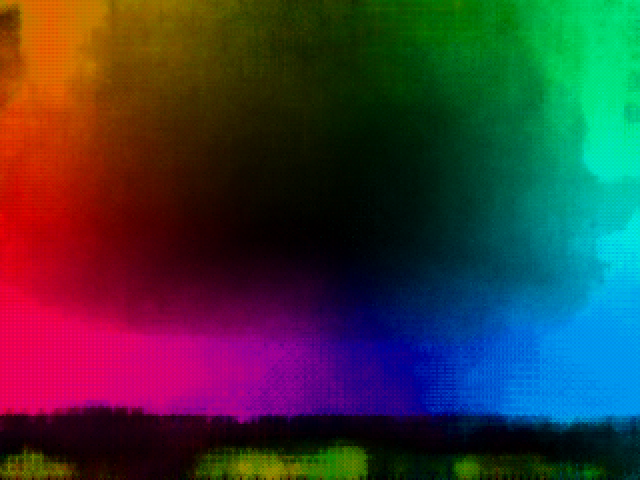}&
\includegraphics[width = 3.4cm]{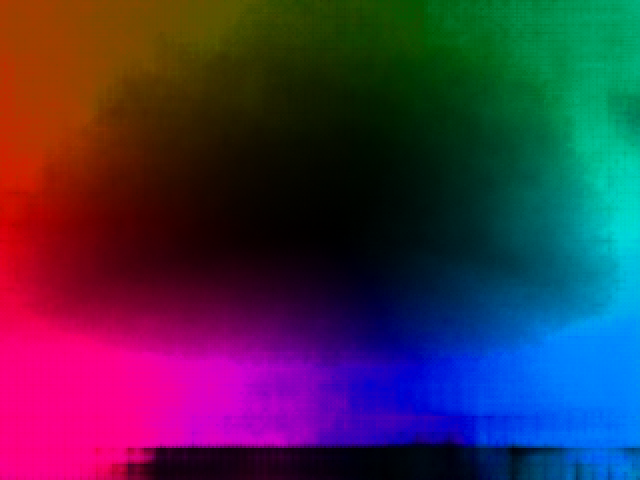}\\

\includegraphics[width = 3.4cm]{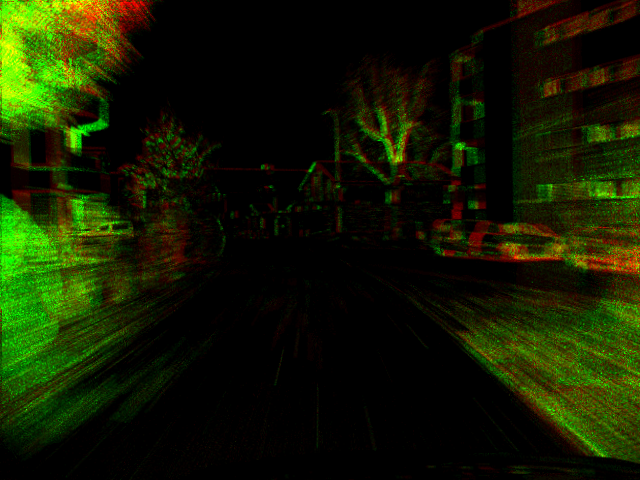}&
\includegraphics[width = 3.4cm]{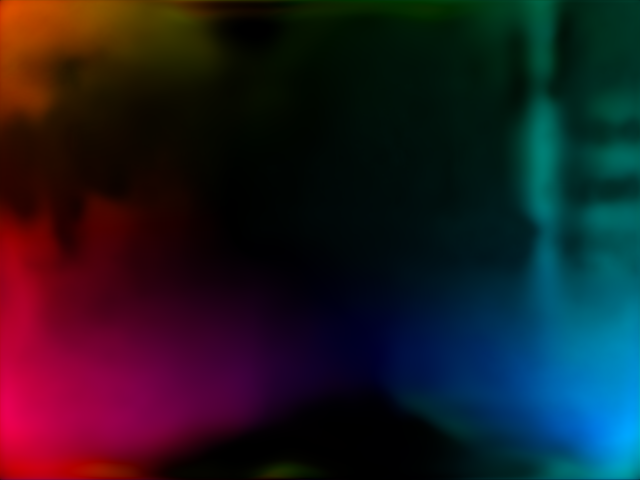}&
\includegraphics[width = 3.4cm]{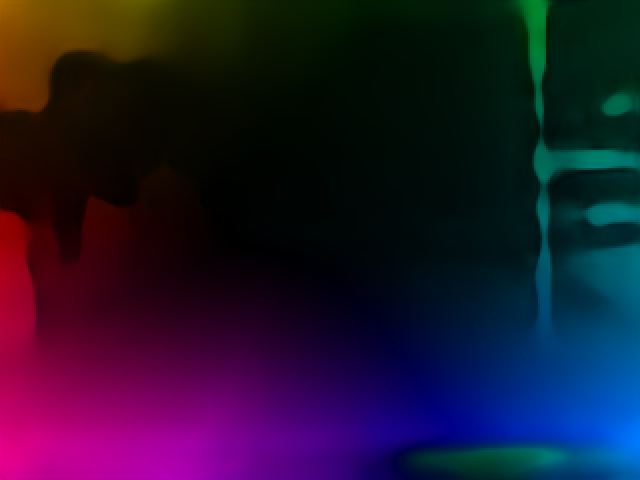}&
\includegraphics[width = 3.4cm]{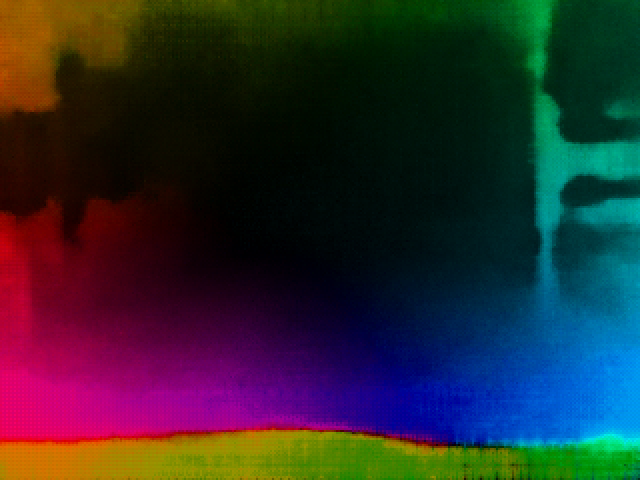}&
\includegraphics[width = 3.4cm]{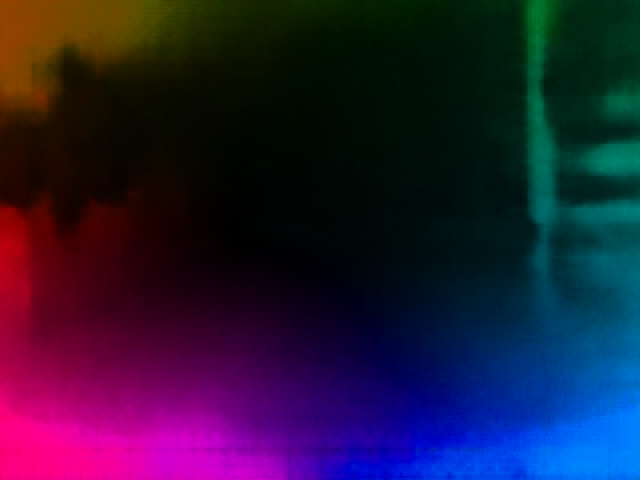}\\
{\footnotesize (a) events} &{\footnotesize (b) EVFlowNet  \cite{zhu2019a}} &{\footnotesize (c) STTFlowNet (ours)} &{\footnotesize (d) SDformerFlow-v1(ours)} &{\footnotesize (e) SDformerFlow-v2(ours)} 
\vspace{0.2cm}
\end{tabular}
\caption{Qualitative results for optical flow evaluated on the official DSEC test subset with resolution $480 \times 640$. The first column presents the event input. Since no ground truth or masks are available. We plot the dense flow estimation for all the models (best viewed in color). }
\label{fig:DSEC_benchmark}
\end{figure*}

Additionally, we evaluate our models on the multi-vehicle stereo event camera dataset (MVSEC)~\cite{zhu2018}. Since the MVSEC and DSEC datasets share different spatial resolutions and ground truth rates, previous works have either trained their work using outdoor day sequences from the same dataset~\cite{zhu2018a,zhu2019a,ding2021} or other small indoor flying dataset~\cite{hagenaars2021a,tian2022etflownet,kosta2023adaptivespikenet}, resulting in overfitting problems \cite{shiba2022a,luo2023ADM-Flow}. To avoid overfitting, we trained our model using the multi-density rendered (MDR) dataset \cite{luo2023ADM-Flow}. MDR was generated using a graphic engine blender and provides 80000 training samples and 6000 validation samples. We trained our network from scratch with cropped resolution $256 \times 256$ and a window size of $2 \times 8 \times 8$ for 50 epochs until convergence. We report our evaluation results for sparse optical flow on the MVSEC dataset to compare with other models.

\subsection{Results}

\subsubsection{Evaluation on the DSEC dataset}
We use average endpoint error (AEE), percentage of outlier over 3 pixels, and average angle error (AAE) as the evaluation metrics. SDformerFlow-v1 uses a LIF model with dot product attention (Fig.~\ref{fig:swinblocks-sdsa}) with 5 time steps while SDformerFlow-v2 uses PSN neurons with SPE embedding and QK attention (Fig.~\ref{fig:swinblocks-qk}) with 10 time steps.

\begin{table}[!t]
  \begin{center}
    \begin{tabular}{ccccc}
      \toprule 
     Training & & AEE & \% Outlier &AAE\\
      \midrule 
      \multirow{5}{*}{A}&
      E-RAFT \citep{Gehrig2021eraft} &0.779&2.684&2.838\\
      &EV-FlowNet\_retrained \citep{Gehrig2021eraft}  &2.320&18.600&-\\
      &IDNet \cite{wu2023IDNet}  &{\bf 0.719}&{\bf 2.036}&2.723\\
       &TMA \cite{liu2023tma}  &0.743&2.301&2.684\\
        &E-Flowformer \cite{li2023blinkflow} &0.759&2.446&{\bf 2.676}\\
        & TamingCM\cite{paredes2023taming} &2.330&17.771&10.560\\
        & STTFlowNet-en3 (Ours) &0.997&4.588&3.235\\
              \midrule
      \multirow{3}{*}{S}
      &OF\_EV\_SNN~\citep{cuadrado2023optical} &1.707&10.308&6.338\\
        & SDformerFlow-v1 (Ours) &2.138&13.967&5.882\\
        & SDformerFlow-v2 (Ours) &\underline{1.602}&\underline{ 10.051}&\underline{4.871}\\
          \midrule
          \multirow{1}{*}{M}
            & MultiCM \cite{shiba2022a} &3.472&30.855&13.983\\

      \bottomrule 
    \end{tabular}
 \vspace{0.2cm}
    \caption{Quantitative results for optical flow estimation on the DSEC optical flow benchmarks for all the test sequences. The first column shows the method type: A stands for ANN, S stands for SNN, and M stands for model-based. We highlight the best-performing results and underline the best among SNN models.}
    \label{tab:DSEC_benchmark_err}
  \end{center}
\end{table}

\begin{table*}[b]
  \begin{center}
    \resizebox{\textwidth}{!}{%
    \begin{tabular}{ccccccccccccc}
      \toprule 
      Training&dt = 1 frame&D&\multicolumn{2}{c}{outdoor\_day1}&\multicolumn{2}{c}{indoor\_flying1}&\multicolumn{2}{c}{indoor\_flying2}&\multicolumn{2}{c}{indoor\_flying3}&\multicolumn{2}{c}{Avg}\\
      \midrule 
      & && AEE & \% Outlier & AEE & \%  Outlier & AEE & \%  Outlier & AEE & \%  Outlier & AEE & \%  Outlier \\
      \midrule 
      \multirow{7}{*}{A}&
      EV-FlowNet \citep{zhu2018a}&M &0.49& 0.20  & 1.03 &2.20& 1.72 & 15.10&1.53 &11.90&1.19&7.35\\
      &EV-FlowNet2 \citep{zhu2019a}&M &\bf{0.32}& \bf{0.00}  & 0.58 &\bf{0.00}& 1.02 & 4.00&0.87 &3.00&0.69&1.75\\
      &GRU-EV-FlowNet \citep{hagenaars2021a} &FPV&0.47& 0.25  & 0.60 &0.51& 1.17 & 8.06&0.93 &5.64&0.79&3.62\\
      &STE-FlowNet \citep{ding2021}&M &0.42& \bf{0.00}  & 0.57 & 0.10&0.79& 1.60 &1.72 &1.30&0.62&0.75\\
        &ET-FlowNet \citep{tian2022etflownet}&FPV &0.39& 0.12  & 0.57 & 0.53&1.20& 8.48 &0.95 &5.73&0.78&3.72\\
        &Adaptive-SpikeNet(ANN)~\cite{kosta2023adaptivespikenet}&M &0.48& - & 0.84 & -&1.59& - &1.36 &-&1.07&-\\
         &ADM-Flow \citep{luo2023ADM-Flow} &MDR&0.41& \bf{0.00}  & \bf{0.52} & 0.14&\bf{0.68} &\bf{1.18} &\bf{0.52}& \bf{0.04}&\bf{0.53}&\bf{0.34}\\
         &STT-FlowNet (ours)&MDR &0.66&0.29  & 0.57 &0.33 &0.88 &4.47 &0.73&1.58&0.71&1.67\\
        \midrule 
       \multirow{5}{*}{S}&
      Spike-FlowNet \citep{lee2020spikeflownet}&M &0.49& - & 0.84 & - &1.28& - &1.11 &-&0.93&-\\
        &XLIF-EV-FlowNet \cite{hagenaars2021a}&FPV &0.45&0.16  & 0.73 &0.92 &1.45&12.18  &1.17 &8.35&0.95&5.40\\
        &Adaptive-SpikeNet \cite{kosta2023adaptivespikenet}&FPV &\underline{0.44}& - & 0.79 & -&1.37&-  &1.11 &-&0.93&- \\
        &Spatiotemporal\_SNN~\cite{zhang2023spatiotemporalsnn}&M  &0.45&\underline{0.00}  &0.76 &\underline{0.00} &1.13&6.00  &0.95 &4.00 &0.82&2.50\\
        &OF\_EV\_SNN \cite{cuadrado2023optical}&M  &0.85&-  & 0.58 &- &\underline{0.72}&-  &\underline{0.67} &- &0.71&-\\
       &SDformerFlow\_v1 (Ours) &MDR & 0.69 &0.21  &0.61  &0.60 &0.83&\underline{3.41} &0.76 &\underline{1.45}&0.72&\underline{1.42}\\
       &SDformerFlow\_{v2} (Ours) &MDR & 0.61 &0.08  &\underline{0.54}  &0.58 &0.81&3.85 &0.69 &1.78&\underline{0.66} &1.57\\

      \bottomrule 
    \end{tabular}
    }
    \caption{Quantitative results for optical flow evaluated on the MVSEC dataset. A and S denote ANN and SNN, respectively. D indicates the training dataset: MVSEC, FPV or MDR. We highlight the best-performing results and underline the best results for the SNN model in each tested sequence. }\label{tab:MVSEC-quantitative}
  \end{center}
\end{table*}

Figure~\ref{fig:DSEC_qualitative} shows qualitative results for both our STTFlowNet and SDformerFlow models, trained with cropped resolution on our split training dataset and tested on the validation dataset. We use bicubic interpolation to remap the relative positional bias for testing on full resolution, as described in~\cite{liu2021swin}. Notably, when the vehicle moves forward in steady motion, all models achieve accurate flow estimation. However, in scenarios involving sharp turns or large, abrupt motions (third row in the figure), the baseline EVFlowNet~\cite{zhu2019a} struggles to estimate the correct direction. In contrast, both our STTFlowNet and our fully spiking model effectively handle such scenarios, thanks to their utilization of spatiotemporal attention mechanisms.

Figure~\ref{fig:DSEC_benchmark} showcases the improved estimation performance of our models on the DSEC optical flow benchmark~\footnote{Full benchmark statistics are available at \href{https://dsec.ifi.uzh.ch/uzh/dsec-flow-optical-flow-benchmark/}{https://dsec.ifi.uzh.ch/uzh/dsec-flow-optical-flow-benchmark/}} test set compared to the baseline. Notably, SDformerFlow-v1 encounters challenges in areas where the sensor hits the car hood for which ground truth data is unavailable. This limitation could be attributed to lack of tuning the LIF neuron statistics. Using PSN with learnable parameters for the SNN neurons alleviates the problem.

\begin{figure*}[t]
\begin{tabular}{c@{\hspace{1mm}}c@{\hspace{1mm}}c@{\hspace{1mm}}c@{\hspace{1mm}}c}
\includegraphics[width = 3.4cm]{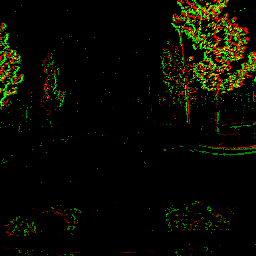}&
\includegraphics[width = 3.4cm]{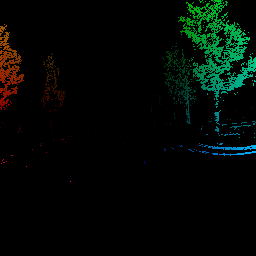}&
\includegraphics[width = 3.4cm]{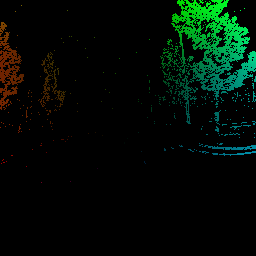}&
\includegraphics[width = 3.4cm]{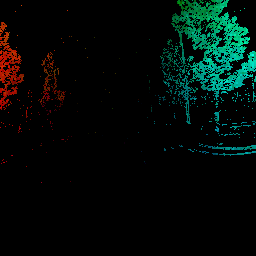}&
\includegraphics[width = 3.4cm]{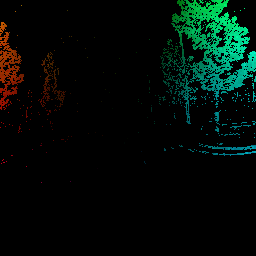}\\
\includegraphics[width = 3.4cm]{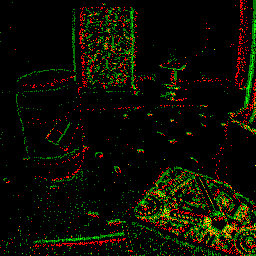}&
\includegraphics[width = 3.4cm]{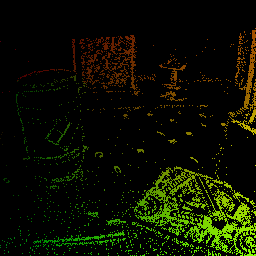}&
\includegraphics[width = 3.4cm]{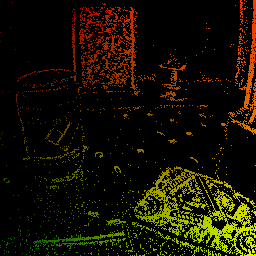}&
\includegraphics[width = 3.4cm]{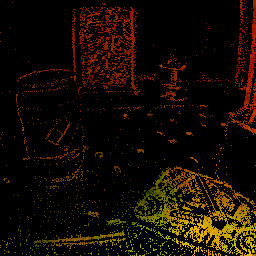}&
\includegraphics[width = 3.4cm]{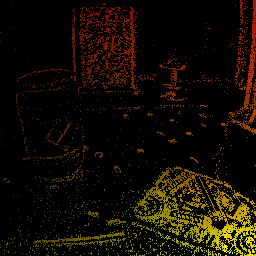}\\

{\footnotesize (a) events} & {\footnotesize (b) ground truth}&{\footnotesize (c) STTFlowNet (Ours)} &{\footnotesize (d) SDformerFlow-v1 (Ours)} &{\footnotesize (e) SDformerFlow-v2 (Ours)} \vspace{0.2cm}
\end{tabular}
\caption{Qualitative results for optical flow evaluated on the MVSEC dataset for the $dt=1$ case. The top row is from the outdoor\_day1 sequence, and the bottom row is from the indoor\_flying1 sequence. The first column presents the event input, and the second column shows the ground truth optical flow provided in the MVSEC dataset. For evaluation, we use the masked sparse optical flow where events are present (best viewed in color).}
\label{fig:MVSEC_qualitative}
\end{figure*}

\begin{table}[t!]
  \begin{center}
    \resizebox{\columnwidth}{!}{%
    \begin{tabular}{ccccccc}
      \toprule 

      Model & EPE& Outlier & AEE & I & Training&Param.\\
        &  & \% &  & & res.  &(M)\\
      \midrule 
     
       EVFlowNet\_retrained &1.63 &10.01 &5.84 &count& 288,384&14.14\\
      EVFlowNet\_retrained &1.57 &9.92 &6.09 &voxel& 288,384&14.14\\
       STTFlowNet-en3-b2-p4-w5 &1.67 &12.61 &8.22 &count&240,320&20.30\\
      STTFlowNet-en3-b2-p2-w10 & 1.34& 8.29&5.98&count&240,320&20.30\\
       STTFlowNet-en3-b4-p4-w10 & 1.37&8.21 & 6.77&count&240,320&20.29\\
        STTFlowNet-en3-b4-p2-w10 &1.43 & 9.44&5.54 &count&240,320&20.29\\
      STTFlowNet-en3-b4-p2-w10& 1.05&4.97 & 5.34&voxel&240,320&20.29\\
  STTFlowNet-en3-b2-p2-w10 &0.94 &3.97 &4.78 &voxel&240,320&20.30\\
         STTFlowNet-en3-b2-p4-w10& 0.83 &2.61 & 4.36&voxel&480,640&20.29\\
          STTFlowNet-en4-b2-p4-w10 & \bf{0.81} & \bf{2.50}&\bf{4.33} &voxel&480,640&57.51\\
      \bottomrule 
    \end{tabular}
    }
   \vspace{0.2cm}
    \caption{Ablation study for STTFlowNet. Column I stands for the event input type. For each algorithm variant, en means the number of encoders, b stands for the number of input blocks, p means spatial patch size, and w stands for swin spatial window size. Best-performing results are highlighted.}
    \label{tab:STT_ablation}
  \end{center}
\end{table}

\begin{table*}[ht]
  \begin{center}
    \resizebox{\textwidth}{!}{%
    \begin{tabular}{ccccccccccccc}
      \toprule 

      Architecture & Shortcut & Encoders&Neuron & \multicolumn{2}{c}{EPE}  & \multicolumn{2}{c}{Outlier \%} & \multicolumn{2}{c}{AEE} & I & Training res. & Param. (M)\\
      \midrule 
        test res: cropped (C) or full (F) &&&& C&F &C &F &C &F &  &  & \\
      \midrule 
    Spiking-EV-FlowNet-s5-c4  &SEW& 4&LIF&3.08&3.47 & 19.67&23.70&17.90&14.41 &10 &288,384&14.13\\

   \cdashline{1-13} 
          SpikeformerFlow-s8-c4  &SEW& 3 &LIF&1.60&3.21 &11.90&32.30&12.51&14.77 &15* &240,320&19.80\\
       SpikeformerFlow-s4-c8 &SEW & 3&LIF&1.76&3.54 &13.43&41.18&14.01&27.81  &15*& 240,320&19.81\\
            SpikeformerFlow-s5-c4 &SEW & 3&LIF&1.51&2.52 &9.85 &22.75& 10.68&11.10 &10 &288,384&19.83\\
               \cdashline{1-13} 
           SDformerFlow-s5-c4    &MS& 3&LIF&1.28&2.01 &6.91&15.55 &9.01&8.99  &10 &288,384&19.83\\

          SDformerFlow-s5-c4   &MS & 4&LIF&1.25&1.98 &6.69&15.06 &8.48&8.81  &10 &288,384&56.48\\
            SpikeCAformerFlow-s5-c4   &MS& 4 &LIF&1.66&2.97 &10.65&27.87 &12.05&22.55  &10 &288,384&15.73\\
            SDformerFlow-s5-c4   &MS & 4&PSN&1.26&1.87 &6.70&13.30 &9.04&9.29  &10 &288,384&56.48\\
   \cdashline{1-13} 
        SDformerFlow-SPE-s5-c4   &MS & 4&LIF&1.17&1.90 &5.45&13.96 &8.03&8.21 &10 &288,384&56.49\\
        SDformerFlow-SPE-QK-s5-c4 &MS & 4&LIF  &1.14&1.96 &4.95&14.12 &7.93&8.38 &10 &288,384&54.92\\
            SDformerFlow-SPE-s5-c4   &MS & 4&PSN&1.08&1.60 &5.23&11.00 &7.21&7.35 &10 &288,384&56.49\\
          SDformerFlow-SPE-QK-s5-c4 &MS  & 4&PSN &1,04&1.64 &4,11&10.80 &7,40&7.66 &10 &288,384&54.92\\
        SDformerFlow-SPE-QK-s10-c2 &MS & 4&PSN &\bf{0.93}&\bf{1.61}
        &\bf{3.17}&\bf{8.91} &\bf{6.37}&\bf{7.23} &10 &288,384&54.92\\

      \bottomrule \\
      \multicolumn{13}{l}{*The SEW variant with input voxel size of 15 was trained with a resolution of $240 \times 320$ due to GPU memory limitations. The rest of the SDformerFlow }\\
      \multicolumn{13}{l}{models were trained at $288 \times 384$ resolution.}
      
    \end{tabular}
    }
   \vspace{0.2cm}
    \caption{Ablation study for SDformerFlow. For the SNN model variants, s stands for number of steps, and c stands for number of channels. SPE stands for the using shortcut for the patch embedding. QK means the model use QK attention while others use dot product attention. Best performing results are highlighted in bold.}
    \label{tab:spikeformer_ablation}
  \end{center}
\end{table*}

Quantitative results for our models, STTFlowNet and SDformerFlow, evaluated on the DSEC benchmark, are presented in Table~\ref{tab:DSEC_benchmark_err}. 
Our ANN model outperforms the baseline model~\cite{zhu2019a} and other self-supervised trained models~\cite{paredes2023taming}. However, it still trails behind correlation-volume-based models~\cite{Gehrig2021eraft,li2023blinkflow}. The only other SNN model included in the benchmark~\cite{cuadrado2023optical} uses stateless neurons and is trained at full resolution, whereas most other SNN approaches are trained and validated on cropped resolution~\cite{ponghiran2023efficient-spikeflow,kosta2023adaptivespikenet} with limited representation in the benchmark. Notably, our fully spiking model, SDformerFlow, exhibits superior performance compared to the ANN baseline~\cite{zhu2019a} and yields state-of-the-art results among all the SNN methods. 

\subsubsection{Evaluation on the MVSEC dataset}
The quantitative evaluation on the MVSEC dataset is given in Table~\ref{tab:MVSEC-quantitative}, and a qualitative evaluation is shown in Fig.~\ref{fig:MVSEC_qualitative}. Both our ANN and SNN models yield competitive results overall. Our models perform slightly below some other methods on the outdoor sequence since these models were trained on the outdoor\_day 2 sequence of the same dataset. On average, our ANN model performs better than other ViT-based U-Net architectures~\cite{tian2022etflownet}. Our improved SDformerFlow-v2 yields state-of-the-art performance for the average error and most indoor sequences among all the SNN methods and is even superior to our ANN model. While the second best performing model~\cite{cuadrado2023optical} reported their results for the indoor sequences separately trained on the subsets of the same dataset, which may have overfitted to the test dataset, our model was trained on a different dataset and shows the generalization capability.

\subsection{Ablation study}

The ablation study was conducted on the validation DSEC set. For the ANN models, we retrained EVFlowNet~\cite{zhu2019a} on the DSEC training set as our base model for 60 epochs while randomly cropping to size $288 \times 384$. Our ANN model shares the same U-Net architecture as EVFlowNet, with the key difference being the use of spatiotemporal swin transformer encoders instead of convolutional layers. Our models were trained at either cropped or full resolution of $480 \times 640$ and validated in full resolution. We analyzed the effects of:
a) the input representation: event voxel or count; b) the number of temporal partitioning blocks: $b2$ or $b4$; c) the spatial patch size: $p=2$ or $p=4$, the swin spatial window size $w$; and d) the training resolution.

Results are summarized in Table~\ref{tab:STT_ablation}. Using the event voxel representation retained more temporal information and notably improved results. The use of swin transformer layers instead of convolutions also led to significant performance gains. For the variants of STTFlowNet, the window size influenced the range of the area to pay attention to, with smaller window sizes making it difficult for the network to learn larger displacements. Adjusting the patch size between 2 and 4 according to the window size and resolution was found to be effective. Partitioning the temporal domain into 2 blocks yielded better results than 4, potentially due to the total number of channels. Further improvements may be achieved by incorporating a local-global chunking approach as described in~\cite{zhang2022MDE}.
To maintain equivalence between our ANN and SNN models, we utilized local temporal blocks exclusively. Given the performance degradation experienced by the swin transformer at higher resolutions, we opted to train the model directly at full resolution using a patch size of 4. This approach ensured that the resolution within the swin encoders remained consistent with training the model at half resolution with a patch size of 2. Notably, this strategy resulted in remarkable improvements in performance.

For our SNN model, we trained the fully spiking version of EVFlowNet~\cite{zhu2019a} with LIF neurons using the same input representations as our base model for comparison. We studied:
a) the number of time steps (s)/number of channels (c); 
b) shortcut variants: SEW or MS shortcuts; 
c) the number of encoders;
d) with or w/o shortcut patch embedding (SPE) module;
e) spiking self-attention variants (dot product or QK); and
f) type of spiking neurons (LIF or PSN).

Results are presented in Table~\ref{tab:spikeformer_ablation}. The spikeformer encoders significantly improved performance compared to the baseline model, albeit with reduced robustness when directly tested on scaled-up resolutions. Incorporating convolution-based modules as CAformer~\cite{yao2024metaformer,yu2024metaformerann} in the first two encoders yielded a lightweight model but with slightly reduced performance. Increasing the number of time steps helped capture temporal information at the expense of increased memory consumption. Opting for 5 time steps and 4 channels struck a balance between performance and memory consumption. The MS shortcut variant notably improved results compared to the SEW shortcut. The possible reason lies in that MS shortcuts provide an information flow path between the states of the neurons before the spike function and are not regulated by their firing status. Increasing the number of encoders from three to four further enhanced performance at the cost of increased parameters. 
Adding the extra shortcut (SPE) for the patch embedding layer notably improved the performance, as the deformed skip connection facilitates the network to learn at the very early stage. Changing the dot product attention into linear QK attention did not hurt the performance, while it reduced the number of parameters to train and memory consumption. Replacing the LIF neurons with PSN neurons did not affect much the performance of the dot product self-attention variant, while it improved the linear QK attention variant. The possible explanation is that the fully spike-driven QK linear attention introduces more sparsity to the model whereas using PSN neurons with learnable parameters facilitates the training process.

\subsection{Energy consumption analysis}

We follow established methodologies from prior research~\cite{yao2024metaformer, ponghiran2023efficient-spikeflow, kosta2023adaptivespikenet} to analyze the theoretical energy consumption for our models. For the ANN models, we estimate energy consumption based on the number of floating-point operations (FLOPS) required. As all operations in ANN layers are multiply-accumulate (MAC) operations, the energy consumption for ANN models is calculated as {\em FLOPS}$\times E_{MAC}$. Conversely, SNN models convert multiplication operations into addition operations due to their binary nature. Thus, for SNN models, we estimate energy consumption by multiplying the FLOPS with the spiking rate $R_s$ and the number of time steps $T$, resulting in {\em FLOPS}$\times R_s \times T \times E_{AC}$. Here, $E_{MAC}$ represents the energy required for MAC operations, and $E_{AC}$ represents the energy required for addition operations. For 32-bit floating-point computation, these energy values are typically $E_{MAC} = 4.6pJ$  and $E_{AC} = 0.9pJ$, respectively, based on 45 nm technology~\cite{Mark2014energy}. We estimate the average spiking rates among all time steps for each layer to calculate energy consumption, ignoring the negligible contribution of batch normalization layers (around 0.01\%). The energy consumption for each model during the inference phase, with an image input size of $288 \times 384$, is presented in Table~\ref{tab:energy_consumption}.
Our results demonstrate that the energy consumption of our SNN model is nearly one-tenth that of its ANN counterpart and one-third that of the baseline EVFlowNet model.

\begin{table}[t!]
  \begin{center}
    \resizebox{\columnwidth}{!}{%
    \begin{tabular}{cccccccc}
      \toprule 

      Model & AEE &Type& Param& FLOPS &Avg. spiking rate&Power\\
             & && (M)& (G) &&(mJ)\\
      \midrule 
     
    EVFlowNet retrained &1.57 &ANN&14.14&22.38&-&102.95\\
    Spiking-EVFlowNet&3.08&SNN&14.13&22.38 &0.29&29.21\\
    STTFlowNet-en3 (ours) &0.72&ANN&20.30&86.88&-&399.65\\
    SDformerFlow-en3 (ours) &1.28&SNN&19.83&45.28&0.27&44.06\\
    SDformerFlow-v1-en4 (ours)&1.25&SNN&56.48&51.27&0.27&48.40\\
    SDformerFlow-v2-en4 (ours)&1.14&SNN&54.92&42.63&0.36&36.83\\

      \bottomrule 
    \end{tabular}
    }
   \vspace{0.2cm}
    \caption{Energy consumption of our ANN and SNN models. en indicates the number of encoders. All the SNN models are based on s5-c4 variant (5 time steps and 4 channels) with LIF neurons. We retrained the EVFlowNet~\cite{zhu2018a} and spiking version of EVFlowNet to be our ANN and SNN baselines.}
    \label{tab:energy_consumption}
  \end{center}
\end{table}

\section{Conclusions}

We introduced STTFlowNet and SDformerFlow, two novel architectures for event-based optical flow estimation that leverage spatiotemporal swin transformer encoders in ANN and SNN frameworks, respectively. Our work marks the first application of using a spikeformer for event-based optical flow estimation. Despite not using correlation volumes and facing scalability challenges inherent to transformer architectures, our results highlight the potential of using spikeformers in regression tasks.
Our ANN model, STTFlowNet, ranks slightly worse in AEE and outlier count with the best-performing volume correlation methods but significantly better than self-supervised ones.
Our SNN version is the first fully spikeformer implementation and yields state-of-the-art performance among the SNN methods on both DSEC and MVSEC benchmarks.   
Notably, our SNN model achieved remarkable energy savings compared to its ANN counterpart and also outperformed the baseline EVFlowNet model.
We believe that by introducing spatiotemporal attention, we strengthen our model’s capability to map global context for the spatial feature maps while capturing spatiotemporal correlations, which improves the performance of our model compared to other CNN-based methods.
However, our model is still based on an encoder-decoder architecture and its performance still falls behind some state-of-the-art ANN methods.
One important limitation of our work is that, by feeding the entire chunk of data into the spatiotemporal attention modules, we are not fully exploiting the asynchronous ability of the event camera and the SNNs. This can be improved in future work by introducing temporal delay, as proposed in~\cite{wang2023STSA}. Thirdly, transformer-based models suffer from constrained scalability across different resolutions. Recent work proposes methods to address this issue by incorporating multi-resolution training~\cite{tian2023resformer} or dynamic resolution adjustment modules~\cite{fan2024vitar}. Finally, much work remains to be done related to hardware implementation to fully exploit the advantage of energy efficiency of SNNs. 
This work highlights the efficacy of integrating transformer architectures with spiking neural networks for efficient and robust optical flow estimation, paving the way for advancements in neuromorphic vision systems.

\bibliographystyle{IEEEtran}
{\small 
\bibliography{bib.bib}  
}

\vfill

\end{document}